 \documentclass[]{elsarticle}
\usepackage[a4paper, total={6in, 8in}, margin=0.9in]{geometry}
\usepackage{multicol}
\usepackage{amssymb}
\usepackage{graphicx}
\usepackage{booktabs}
\usepackage{multirow}
 
\usepackage{amsmath}
\usepackage{caption}
\usepackage{subcaption}
\usepackage{algorithm}
\usepackage{algpseudocode}

\usepackage{amsmath,amssymb,amsfonts}
\usepackage{algorithm}
\usepackage{algpseudocode}
\usepackage{graphicx}
\usepackage{textcomp}
\usepackage{multirow}
\usepackage{longtable}
\usepackage{rotating} 
\usepackage{longtable}
\usepackage{booktabs}
\usepackage{ltxtable} 
\usepackage{supertabular,booktabs}
\usepackage[export]{adjustbox}
\usepackage{xcolor}

\usepackage{changes}

\begin{document}

\begin{frontmatter}



\title{Patient Similarity Computation for Clinical Decision Support: An Efficient Use of Data Transformation, Combining Static and Time Series Data}

%% AUTHORS %%%%%%%%%%%%%%%%%%%%%%%%%%%%%%%%%%%%%%%%%%%%%%%%%%%%%%%%%%%%%%%%%%%%
%% Leave this section commented out so that the paper is blinded for review.

%% AUTHORS %%%%%%%%%%%%%%%%%%%%%%%%%%%%%%%%%%%%%%%%%%%%%%%%%%%%%%%%%%%%%%%%%%%%
%% Leave this section commented out so that the paper is blinded for review.
\author[ss]{Joydeb Kumar Sana}
 \address[ss]{Department of Computer Science and Engineering, Bangladesh University of Engineering and Technology, Dhaka, Bangladesh (e-mail: joydebsana@iict.buet.ac.bd, joysana@gmail.com)}
  
% \fntext[ssnote]{Anything you want to add as a footnote about Susan.}

\author[bb]{Mohammad M. Masud}
  \address[bb]{College of Information Technology, United Arab Emirates University (e-mail: m.masud@uaeu.ac.ae)}

%\author[cc]{Muhammad Abdullah Adnan}
%  \address[cc]{Department of Computer Science and Engineering, Bangladesh University of Engineering and Technology, Dhaka, Bangladesh (e-mail: adnan@cse.buet.ac.bd)}
  
\author[cc]{M Sohel Rahman}
  \address[cc]{Department of Computer Science and Engineering, Bangladesh University of Engineering and Technology, Dhaka, Bangladesh (e-mail: msrahman@cse.buet.ac.bd, sohel.kcl@gmail.com)}

\author[dd]{M Saifur Rahman}
 \address[dd]{Department of Computer Science and Engineering, Bangladesh University of Engineering and Technology, Dhaka, Bangladesh (e-mail: mrahman@cse.buet.ac.bd)}

%% Only give the email address of the corresponding author
 \cortext[cor]{M Saifur Rahman}
 \ead{mrahman@cse.buet.ac.bd}
 
%%%%%%%%%%%%%%%%%%%%%%%%%%%%%%%%%%%%%%%%%%%%%%%%%%%%%%%%%%%%%%%%%%%%%%%%%%%%%%%%


%\author{}

%\affiliation{organization={},%Department and Organization
 %           city={},
  %          postcode={}, 
   %         state={},
    %        country={}}

\author{  }

\begin{abstract}
Patient similarity computation (PSC) is a fundamental problem in healthcare informatics. The aim of the patient similarity computation is to measure the similarity among patients according to their historical clinical records, which helps to improve clinical decision support. This paper presents a novel distributed patient similarity computation (DPSC) technique based on data transformation (DT) methods, utilizing an effective combination of time series and static data. Time series data are sensor-collected patients' information, including metrics like heart rate, blood pressure, Oxygen saturation, respiration, etc. The static data are mainly patient background and demographic data, including age, weight, height, gender, etc. Static data has been used for clustering the patients. Before feeding the static data to the machine learning model adaptive Weight-of-Evidence (aWOE) and Z-score data transformation (DT) methods have been performed, which improve the prediction performances. In aWOE-based patient similarity models, sensitive patient information has been processed using aWOE which preserves the data privacy of the trained models. We used the Dynamic Time Warping (DTW) approach, which is robust and very popular, for time series similarity. However, DTW is not suitable for big data due to the significant computational run-time. To overcome this problem, distributed DTW computation is used in this study. For Coronary Artery Disease, our DT based approach boosts prediction performance by as much as 11.4\%, 10.20\%, and 12.6\% in terms of AUC, accuracy, and F-measure, respectively. In the case of Congestive Heart Failure (CHF), our proposed method achieves performance enhancement up to 15.9\%,10.5\%, and 21.9\% for the same measures, respectively. The proposed method reduces the computation time by as high as 40\%. We are confident in the sturdiness and effectiveness of the proposed technique for finding similar patients, making it well-suited for integration into clinical decision support systems.
\end{abstract}

\begin{keyword}
Patient similarity \sep  Data transformation \sep Spark computation \sep Machine learning \sep Time series  \sep Dynamic time warping \sep Health informatics  \sep Distributed patient similarity \sep Data privacy
\end{keyword}

\end{frontmatter}

\section{Introduction}
In recent years, significant focus has been directed towards research on clinical decision support, owing to the development of digital healthcare systems and the availability of huge medical data sourced from various outlets, including electronic health records (EHR), medical imaging, clinical tests, treatment records, genetic information, public health data, etc. Researchers are using those diverse sorts of patients' personal and clinical information to acquire particular goals. These clinical data can be used for patient similarity computation to support clinical decisions within a minimum amount of time and may save many patients' lives \cite{Yi_Mao_2012}. It also enhances clinical decision making without incurring extra effort from physicians. Patient similarity serves as a cornerstone for numerous healthcare applications, including cohort analysis, case-based reasoning, treatment comparisons, disease sub-typing, personalized medicine, etc. \cite{Parimbelli_2018} \cite{Suo_2018}. Moreover, patient similarity prediction is a fundamental problem in evidence-based medicine, recognized as a pivotal area for reshaping healthcare and enhancing the quality of care delivery \cite{Zhu_2016}.
 
While the concept of patient similarity is not new, and several similarity prediction approaches are available~\cite{Zhang2022, ZhengJia2020, Zhu2016, Sanjay2018}, the prediction results of these methods are unsatisfactory and have not been widely implemented. Patient similarity techniques have recently demonstrated success in predicting cancer and supporting various clinical decision systems \cite{Pai_2019, Wang_2014}. To validate the effectiveness of our patient similarity model, we applied it to the prediction of Coronary Artery Disease (CAD) and Congestive Heart Failure (CHF), two prevalent and life-threatening Cardiovascular diseases (CVDs) (also known as heart disease). While various studies have explored patient similarity for predicting outcomes such as hospital length of stay, mortality, trauma, and hospital readmission, further investigation into its application for cardiovascular disease prediction could be beneficial. Coronary artery disease is a disease of the main blood vessels that supply blood to the heart. Congestive Heart Failure, on the other hand, is a heart condition in which the heart cannot pump sufficient blood around the body. This may be due to the heart not filling with enough blood or it being too weak to pump correctly. It refers to the inadequate functioning of the heart muscle, causing fluid to accumulate in the lungs, abdomen, feet, and arms. Hence, the term ``congestive'' is used. According to the World Health Organization (WHO), CVD is the most life threatening disease and many patients experience CVD symptoms that were previously overlooked or unrecognized before their critical condition or death. By leveraging patient similarity techniques, our model aims to identify individuals at high risk for CAD and CHF based on historical patient data, enabling earlier diagnosis and intervention. This validation demonstrates how patient similarity can improve disease prediction beyond traditional risk assessment models, ultimately contributing to more personalized and effective clinical decision-making.

Upon a patient's arrival at the hospital for emergency treatment, their observable vital symptoms, continuously collected time series readings from sensors, and their historical and demographic information can be leveraged to identify the most similar patient within the database. With the help of the medical records of the similar patient, the medical practitioner can reach a more informed and confident decisions regarding the diagnosis, treatment, medication, hospital admission, and other aspects concerning the patient's care. Thus, identifying patient similarity through the amalgamation of time series and static data can be a valuable asset, particularly in situations where other clinical data types (such as lab tests) are not instantly at hand. In this research, both time series and static data have been used to predict Coronary artery disease (CAD) and Congestive Heart Failure (CHF).

\subsection{Literature Review} \label{sec:literature_review} 
Clinical decision support has been discussed in the literature using various techniques including machine learning (ML), data mining, time series similarity computation techniques, patient similarity network approaches, etc. Various machine learning (ML) and data mining techniques such as Bayesian Network \cite{Cai2015}, Support Vector Machine (SVM) \cite{Chan_2010}, Local Spline Regression (LSR) \cite{F_Wang_2012}, CNN \cite{Yu_Cheng_2016} \cite{Che2017ExploitingCN} have been proposed to predict the patient similarity using electronic health record data. An extensive survey on data mining applications in health informatics and big data was performed by Herland et al. \cite{Herland2013}. Another survey on patient similarity was presented by Anis et al. \cite{Anis2017}. In \cite{Evan2022}, Evan et al. predicted the risk for trauma patients using gradient boosting classifier on MIMIC-III database, where the authors used various learning algorithms. Lei et al. \cite{LeiLin2019} proposed a time-series based classification in the field of medical diagnosis. In \cite{Sanjay2018}, Purushotham et al. utilized both static and time series features from the MIMIC-III clinical care dataset to construct a deep learning model to predict the mortality rate, length of stay, and ICD-9 diagnosis code group (e.g. respiratory system diagnosis). In \cite{Ng2015PersonalizedPM}, Ng et al. demonstrated a personalized predictive healthcare model by matching clinically similar patients through a locally supervised metric learning approach. Park et al. \cite{Park2006} investigated the optimum number of neighbors for each patient based on the distribution of pairwise distances. David et al. \cite{David2011} employed the Euclidean distance on weighted predictors to identify neighbors for a given patient. Hielscher et al. \cite{Hielscher2014} proposed the idea of subgrouping the training set based on gender. Then applying a KNN algorithm they showed that it can reduce the dimension of the predictor space and improve prediction performance. Sun et al. \cite{Sun2010} employed a linear regression model based on a least squared error fitting technique. Wang et al. \cite{Wang2011} presented a multiple patient similarity models learned independently on a private dataset. In another work, Wang et al. \cite{Wang20212} incorporated expert knowledge by leveraging two matrices: a similarity matrix and a dissimilarity matrix. Lowsky et al. \cite{Lowsky2013} proposed a neighborhood-based survival probability prediction model based on a Mahalanobis distance. In \cite{Kasabov2010IntegratedOM}, Nikola et al. introduced a integrated method for Personalized Modelling (IMPM) to facilitate personalized treatment and personalized drug design. Besides those, clustering methods have also been adopted in many studies to calculate patient similarity. To assess patients' health perceptions, Sewitch \cite{Sewitch2004} utilized K-means cluster analysis to identify patient groups based on multivariate patterns. To capture the underlying structure in the history of present illness, Henao et al. \cite{Henao2013} presented a clustering model that groups patients based on the historical data of present illness. Huang \cite{Huang2013} proposed a recursive K-means spectral clustering method (ReKS) for disease gene expression data to analyze human diseases.

Most of these studies have demonstrated the effectiveness of their models from different perspectives. However, the performance of patient similarity models is not remarkably satisfactory for real-world use. Moreover, the previous studies did not work on Congestive Heart Failure for patient similarity. Therefore, there is significant room to explore new techniques to improve the performance of patient similarity computation. In this research, we utilized both time series and static data of patients. Finding similar patterns and behaviors in time series data is an interesting research topic with numerous real-world applications. To calculate the distance or similarity between the two time series data, Dynamic Time Warping (DTW) algorithm \cite{Donald_1994} has been used. DTW is the most popular, robust and well-established method for time series similarity and successfully implemented in various applications \cite{Vaughan_2016}, \cite{Franses_2020},\cite{brian2020}, \cite{Donald_1994}. The utilization of DTW in conjunction with the nearest neighbor rule has proven effective in numerous time series classifier tasks \cite{brian2020}. A straightforward nearest neighbor recognition approach combined with DTW is more accurate compared to various advanced classifiers, such as decision trees, artificial neural networks, Bayesian networks, Support vector machines, etc. \cite{Abdelmadjid2021}. However, the major drawback of DTW is its time complexity, especially for large datasets featuring lengthy sequences \cite{brian2020}. It may be impossible to train a model within a reasonable time frame. Moreover, DTW involves significant computation resources \cite{Mehedy_Masud_2020}. To tackle this issue, in this paper, we investigate distributed patient similarity computation using Spark \cite{spark_1} to compute DTW based time series similarities. Recognizing that a basic nearest neighbor strategy combined with DTW achieves higher performance, we propose a distributed model to compute the patient's nearest neighborhood utilizing the time series data aggregated with the static data. We utilized multivariate series to compute the DTW based similarity. The nearest neighbor approach used in this study is based on the neighborhood population (NPop) fusion, as mentioned in \cite{Mehedy_Masud_2020}. NPop is very similar to Affinity network Fusion (AFN) presented by Ma et al. \cite{Ma_2017}, for cancer patient clustering. AFN is the modified version of Patient Similarity Networks (PSNs) presented in \cite{Pai_2018} and \cite{Wang_2014}, which converts heterogeneous clinical data into usable comprehensive network views that are beneficial for clinical decision support. Nearest neighborhood fusion is generated from the outcome of DTW based time series similarity and static data based clustering. On static data, we performed four different clustering algorithms: spectral clustering, K-means clustering,  Agglomerative clustering, and OPTICS (Ordering Points To Identify the Clustering Structure). Before feeding the static data to the clustering algorithms, we experimented with the application of two data transformation (DT) methods, namely, adaptive Weight-of-evidence (aWOE) and Z-score, on the data. The aWOE DT method is a modification of Weight-of-evidence (WOE) developed in our previous research \cite{sana2024ppccp} and it shows remarkable performance improvement. Another study \cite{joydebSana2022} reported that the Z-score notably enhances prediction performance. Those evidences motivate us to test the effect of those two data transformation methods in the field of patient similarity prediction (PSP). Extensive experiments have been conducted on MIMIC-III datasets \ref{MIMICIII} to evaluate the performance of the proposed distributed patient similarity computation framework. 

\subsection{Our Contributions} \label{sec:our_contribution} 
Developing a PSP system poses several challenges, particularly in the context of similarity models, which are more intricate compared to conventional classification algorithms. Firstly, time series data collected from different patients might differ in sampling frequency and sample counts. Secondly, certain patient records may lack specific time series data, leading to missing values. Lastly, integrating time series data with static data poses a notable challenge, which is commonly encountered in medical datasets. Our work focuses on tackling these challenges. In particular, this paper has the following key contributions: 
\begin{itemize}
    \item This study leverages an effective integration of time series and static data. Dynamic Time Warping (DTW), known for its robustness in time series similarity computations, was employed for patient similarity computation. Distributed Spark computation has been used for patient similarity prediction. Our experimental results show that distributed Spark computation notably reduces computation time. To the best of our knowledge, this proposed method is the first implementation of distributed Spark and DTW based time series similarity computation in conjunction with data transformation based clustering techniques for patient similarity prediction. 

    \item Four clustering methods, including Spectral, \textit{K}-Means, Agglomerative, and OPTICS (Ordering Points To Identify the Clustering Structure), have been explored to identify the nearest neighboring patient in conjunction with a time series similarity approach for a targeted patient.

    \item Two DT methods have been examined in the context of PSP system: adaptive Weight-of-Evidence (aWOE) and Z-score. Eventually, to preserve data privacy, aWOE has been applied on the patients' demographic and medical records.

    \item We worked on two prediction problems in this research: Coronary Artery Disease (CAD) and Congestive Heart Failure (CHF). The proposed methodology has been assessed using six evaluation metrics: AUC, accuracy, specificity, precision, recall, and F-measure. The models have been compared against several baseline models, as well as a state-of-the-art technique. The developed methodology notably improves the prediction performance for both the prediction tasks.
    
    %\item To preserve data privacy, aWOE has been applied on the patients' demographic and medical records.
    %\item We conducted Silhouette analysis to determine the appropriate number of clusters. However, the combined results don't align with the output of the Silhouette analysis.
    %\item The proposed methodology has been assessed using six evaluation metrics: AUC, accuracy, specificity, precision, recall, and F-measure. The models have been compared against several baseline models, as well as a state-of-the-art technique. The developed methodology notably improves the prediction performance for both prediction problems (Coronary Artery Disease and Congestive Heart Failure).
      
    %\item Statistical significance tests were conducted on our experimental findings, and the results clearly indicate that the performance of the proposed method was statistically significant.
\end{itemize}

The rest of the paper is organized as follows: Section \ref{sec:methodology} describes the methodology and framework of the proposed study. The experimental results and comparisons are briefly explained in Section \ref{sec:experimental_result}. Statistical significance analysis is presented in Section \ref{sec:performance_analysis}. Section \ref{sec:comparison_previous_study} provides a comparison among the proposed models, previous studies, and baseline models. The impact of adaptive Weight-of-Evidence on model output is discussed in Section \ref{sec:impact_aWOE}. Section \ref{sec:aWOE_privacy} addresses the privacy assessment. The discussion is presented in Section \ref{sec:discussion}. We conclude the proposed study in Section \ref{sec:conclusions}

\section{Methodology} \label{sec:methodology} 

\begin{figure}[!htb]
\begin{center}
\includegraphics[height=150px,width=480px]{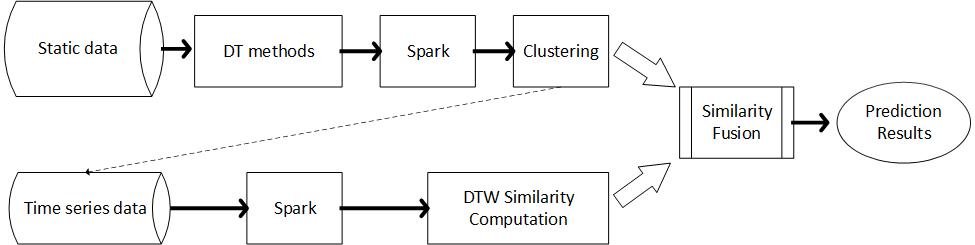}
\caption{High level block diagram of DT based distributed patient similarity model (DT-DPSM)}
\label{fig:DPSM_framework}
\end{center}
\end{figure}

In this section, we provide a detailed description of the proposed empirical study, experimental setup, and evaluation approaches. A high level block diagram of our proposed patient similarity model is shown in Figure \ref{fig:DPSM_framework}. The model used both time series data and static data. Firstly, patients' demographic and medical history data (static data) have been utilized to cluster them into groups. Prior to clustering, we performed two data transformation methods on the static data. The clustering process has been conducted in Spark environment. Secondly, based on the group of the targeted patients, a pair-wise patient similarity matrix has been calculated using the dynamic time warping (DTW) method for each time series variate. Distributed Spark system has been used to calculate the similarity matrix. Thirdly, a similarity fusion has been generated by combining the similarity matrices. Finally, the nearest similar patient is selected from the similarity fusion. In the following part of this section, we will discuss each of these steps individually.
%Firstly, pair wise patient similarity matrix has been calculated using dynamic time warping (DTW) method for each time series variate. Distributed Spark system has been used to calculate the similarity matrix. Secondly, patients demographic and medical history data (static data) have been utilized to cluster them into groups. Before perform clustering, we performed two data transformation methods on the static data. The clustering process have been performed using spark computation. Thirdly, Combine with the similarity matrices and based on the clusters a similarity fusion matrix has been generated. Finally, nearest similarity patient is selected from the fusion matrix based on the majority voting. In the following part of this section, we will discuss each of these steps individually.

\subsection{Datasets and Data Preprocessing} \label{sec:Datasets} 
In this study, MIMIC-III \cite{MIMICIII} dataset was used for the experiments. The database was collected by MIT's Laboratory of Computational Physiology as part of the Multi-parameter Intelligent Monitoring in Intensive Care (MIMIC) project, financially supported by the National Institute of Biomedical Imaging and Bioengineering. It is a large, freely accessible database containing de-identified health data associated with more than forty thousand patients who were admitted to critical care units at the Beth Israel Deaconess Medical Center from 2001 to 2012. MIMIC-III includes information associated with 53,423 separate hospital admissions involving adult patients aged 15 years or above, along with 7,870 neonates admitted to the BIDMC. The dataset involves 38,597 unique adult patients across 49,785 hospital admissions. Out of the total admissions, we have specifically chosen data concerning cardiovascular-related patients (all types of diseases that affect the heart or blood vessels, including coronary heart disease, heart attacks, stroke, heart failure, heart block, peripheral artery disease, congestive heart failure, etc.) as our focus lies on computing the similarity among individuals within this category. The total count of selected instances amounts to 4,418. Following a successful completion of a web-based training course (part-1 and part-2) of the National Institutes of Health Protecting Human Research Participants, we achieved the authorization to access the data from MIMIC-III for research objectives (Certification Record ID: 47225752). 
 
 \begin{longtable}[c]{  p{3cm}  p{7cm} p{3cm}  }
 \caption{List of Time series items.\label{table:TimeSriesData}}\\
%\label{table:DTMethods}
 \hline
 \multicolumn{3}{ c }{Begin of Table}\\
 \hline
ItemID & Description &  No. of Non Empty Patients\\
 \hline
 \endfirsthead

 \hline
 \multicolumn{3}{c}{Continuation of Table \ref{table:TimeSriesData}}\\
 \hline
ItemID & Description & No. of Non Empty Patients\\
 \hline
 \endhead

 \hline
 \endfoot

 \hline
 \multicolumn{3}{ c }{End of Table}\\
 \hline
 \endlastfoot

 \vspace{.05mm}\\
1529 & Glucose & 2410 \\
220045 & Heart Rate & 2022 \\
220047 & Heart Rate Alarm Low & 1675 \\
220050 & Arterial Blood Pressure Systolic & 1412 \\
220051 & Arterial Blood Pressure diastolic & 1412 \\
220052 & Arterial Blood Presssure Mean  & 1416 \\
220059 & Pulmonary artery pressure systolic & 970 \\
220060 & Pulmonary artery pressure diastolic & 970 \\
220061 & Pulmonary artery pressure mean & 1020 \\
220074 & Central venous pressure & 1314 \\
220210 & Respiratory Rate & 2020 \\
223761 & Body Temperature & 1962 \\
223834 & O2 Flow & 1920 \\
224161 & Resp Alarm High & 1995 \\
%224639 & Daily weight & 1462 \\
224687 & Minute volume & 1481 \\
224688 & Respiratory Rate (Set) & 1372 \\
224695 & Peak insp. Pressure & 1470 \\
224697 & Mean Airway Pressure & 1475 \\
\end{longtable}

\subsubsection{Static Data} \label{sec:static_data}
We utilize various demographic features as static data, including age, weight, height, and gender. Several background features such as  Admission type, Coronary Artery Disease (not employed for Coronary Artery Disease prediction), and Congestive Heart Failure (not employed for Congestive Heart Failure prediction) have also been incorporated. We use these static data for clustering. For the prediction evaluation, we used Coronary Artery Disease, and Congestive Heart Failure. Coronary Artery Disease, Admission type, and Congestive Heart Failure were categorical but have been transformed into binary format. In the case of Coronary Artery Disease,  '1' signifies 'coronary artery disease,' while '0' represents other Diagnoses. Congestive Heart Failure has been binary-encoded, with '1' representing Congestive Heart Failure and '0' for other types.  Data transformation methods have been applied on the static data prior to utilization.

\subsubsection{Time Series Data}\label{sec:time_series_data}
\begin{figure}[!htb]
\begin{center}
\includegraphics[height=275px,width=520px]{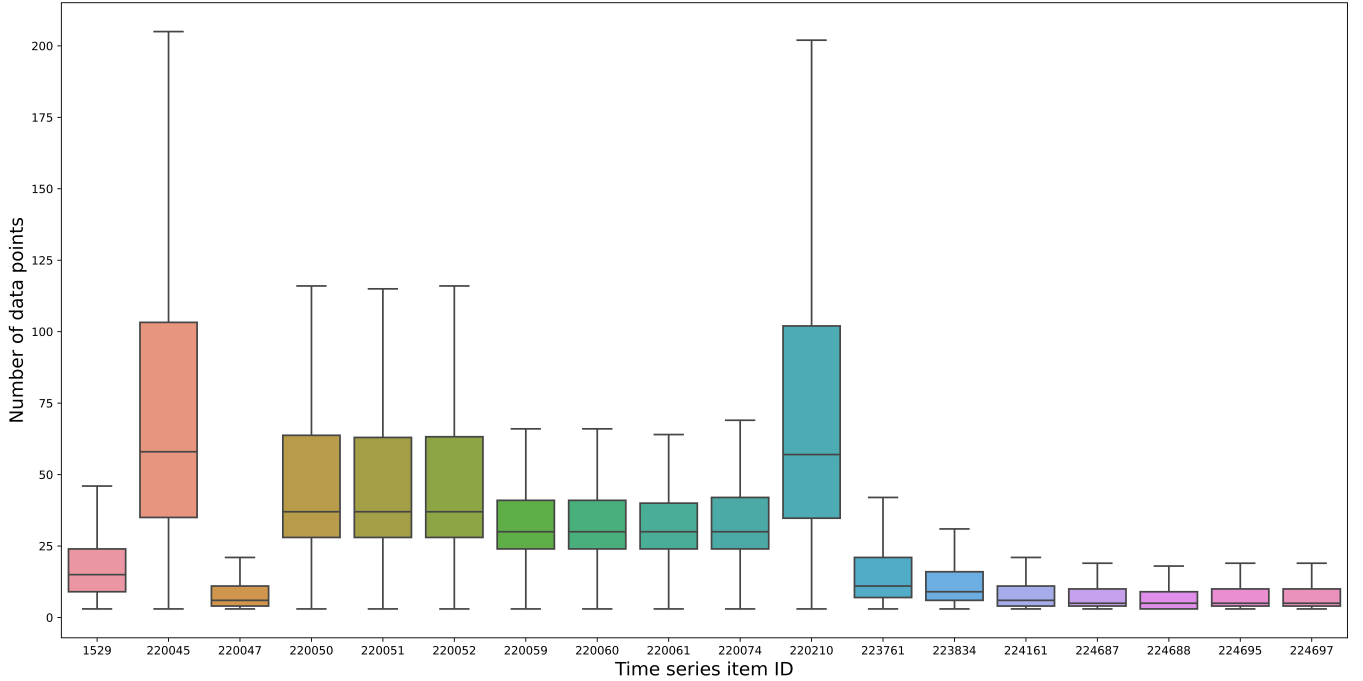}
\caption{The box plot diagram shows the distribution of the number of data points in various time series data items, excluding outliers.}
\label{fig:Box-plot-time-series}
\end{center}
\end{figure}

In this research, 18 different time series data presented in Table \ref{table:TimeSriesData} have been chosen for our experiments. The third column in the table denotes the count of non-empty patient records among the 4,418 instances, captured for the particular time series data. These data are extracted from Chartevents table within MIMIC-III dataset. The majority of the samples are collected at an hourly rate. The box plot diagram in Figure \ref{fig:Box-plot-time-series} displays the distribution of the number of data points across time series data.

The dataset is available  at the following link:\\
 {\color{blue}{https://physionet.org/content/mimiciii/1.4/ (Last Access: November 22, 2021)}}.

\subsection{Data Transformation}\label{sec:DT}
Data transformation is the technique of converting one data format to another format, aiding the data mining process to identify useful data patterns that contribute to decision-making systems. Various data transformation methods have been used in different research fields like Telecommunication industry \cite{joydebSana2022}, E-commerce sector \cite{Satu2023}, etc. Our previous studies \cite{joydebSana2022} and \cite{sana2024ppccp} show aWOE and Z-score data transformations have a positive impact on improving prediction performance. We applied those two data transformation methods on the static data before using it for clustering.

\subsubsection{Adaptive Weight-of-evidence (aWOE)}\label{sec:woe} 
The Weight-of-evidence (WOE) \cite{joydebSana2022}, \cite{COUSSEMENT201727} indicates how influential an independent variable is concerning the dependent variable's prediction. It represents the relationship between a predictive variable and a binary outcome. This conversion makes it easier to understand how various variables affect the desired outcome, particularly in situations involving numerous categorical or numerical variables. A high WOE value indicates a stronger relationship between the feature category and the outcome, while a low WOE value indicates a weaker connection. In most cases, WOE addresses the skewness in the data distribution \cite{joydebSana2022},\cite{COUSSEMENT201727}. The term WOE is calculated as the natural logarithm (ln) of the ratio between the distribution of events (1) and the distribution of non-events (0) in a particular bin. Adaptive Weight-of-evidence (aWOE) \cite{sana2024ppccp} is an extension of WOE. The number of bins of aWOE is determined by dividing the sample size by $q$, when the number of unique values of the feature is greater than $\lambda  ~(\lambda=100)$. Otherwise, the number of bins is equal to the number of unique values of the feature. aWOE also uses an adjustment constant $\epsilon~(\epsilon = 0.0001)$ in both the numerator and denominator. The mathematical equation of the aWOE can be expressed as:

\begin{equation} \label{eq:a_Weightofevidence}
       aWOE = ln\bigg(\frac{\text{Distribution of positive events in a particular bin +  $\epsilon$ }}{\text{Distribution of negative events in a particular bin + $\epsilon$ }}\bigg)
\end{equation}

\subsubsection{Z-score}\label{sec:zscore}
 The procedure of this DT method is useful for normalizing and comparing data from different distributions, allowing for a better understanding and comparison of data points within a dataset \cite{Cheadle_2003}. It represents the distance of a data point from the mean in terms of standard deviation units. It is calculated by subtracting the mean of the dataset from the individual data point and then dividing that result by the standard deviation of the dataset. The formula is given below. 
 \begin{equation} \label{eq:zscore_transformation}
     \text{Z-score} =\frac{x - \text{sample mean}}{\text{sample standard deviation}}
\end{equation}
where $x$ denotes a specific value of any feature within the original dataset.

\subsection{Clustering with Static Data} \label{sec:clustering}
%From time series datasets we calculate $M_\mathcal{F}$ which is a combination for K different distance matrices. To find the nearest neighbor for each matrix in $M_\mathcal{F}$ for a target patient we partition the patients into different groups (clusters).  
To generate the clusters, we utilized the static data of the patients, encompassing factors like age, weight, height, and gender. The historical clinical information of patients, such as Coronary Artery Disease, Admission type, and Congestive Heart Failure, are also utilized in clustering. When focusing on the Coronary Artery Disease, we consider other features while disregarding the Coronary Artery Disease attribute. For targeting Congestive Heart Failure, Coronary Artery Disease, and additional features are considered, while Congestive Heart Failure itself is disregarded. Four clustering algorithms, namely Spectral clustering \cite{Yu_Shi_2003}, K-Means clustering \cite{Kanungo_2002}, Agglomerative clustering \cite{Zepeda_Mendoza2013}, and OPTICS (Ordering Points To Identify the Clustering Structure) \cite{Ankerst_1999} have been used for clustering the patients. Table \ref{table:ClusteringMethods} provides a description of the clustering methods leveraged in our research. The first three algorithms require the user to provide the input parameter $K$, which represents the number of clusters or groups to generate. The clustering algorithms partition the patients into $K$ number of groups. The fourth algorithm (OPTICS) also partitions the patients into clusters but does not require the user to specify the number of clusters. Instead, it requires a minimum sample size. For any target patient, its neighboring patient is determined solely within its own cluster.

\begin{longtable}[c]{  p{3cm}  p{10cm}  }
 \caption{List of Clustering Methods.}\\
 \label{table:ClusteringMethods}\\
 \hline
 \multicolumn{2}{ c }{Begin of Table}\\
 \hline
  Clustering Method & Description \\
 \hline
 \endfirsthead

 \hline
 \multicolumn{2}{c}{Continuation of Table \ref{table:ClusteringMethods}}\\
 \hline
  Clustering Method & Description \\
 \hline
 \endhead

 \hline
 \endfoot

 \hline
 \multicolumn{2}{ c }{End of Table}\\
 \hline\hline
 \endlastfoot

 \hline
 \vspace{.05mm}\\

 Spectral clustering  & 
  Spectral clustering is a graph theory based approach used to identify communities of vertices in a graph based on the edges connecting them. This approach is flexible and enables clustering of non-graph data as well, either with or without the original data. This method is easy to implement and reasonably fast, particularly on sparse datasets comprising up to several thousand entries. It clusters the data points as a graph partitioning problem without presuming the shape or structure of the data clusters. 
 \\
  \vspace{.01mm}\\  
 K-Means clustering & 
 The K-Means clustering algorithm partitions a dataset into $K$ distinct non-overlapping clusters or subgroups, assigning each data point exclusively to a single group. The aim of the algorithm is to make as similar as possible among data points within clusters while ensuring maximal dissimilarity (distance) among the clusters themselves. It chooses data points within a cluster to minimize the sum of the squared distance between these points and the centroid of the cluster.

 \\
  \vspace{.01mm}\\
 Agglomerative clustering &  
Agglomerative clustering is a hierarchical clustering technique commonly used in data mining and machine learning. It follows a bottom-up approach, where each data point initially forms its own cluster. Clusters are then iteratively merged based on similarity, determined using a specified distance metric such as Euclidean distance. This process continues until either all data points are combined into a single cluster or a predefined stopping criterion, such as a specified number of clusters ($K$), is met \cite{Zepeda_Mendoza2013}.
 \\
 \vspace{.01mm}\\

 OPTICS clustering &   
The OPTICS (Ordering Points To Identify the Clustering Structure) clustering algorithm identifies clusters based on the density of points in the dataset. It defines clusters as areas where there is a high density of data points, separated by regions with lower density \cite{Ankerst1999}. It is an extension of the DBSCAN (Density-Based Spatial Clustering of Applications with Noise) clustering method. The main disadvantage of DBSCAN is that it struggles to identify clusters in data that poses a variety of density. OPTICS doesn’t require the density to be consistent across the dataset. OPTICS doesn't require specifying the maximum distance between points (epsilon) as in DBSCAN. However, it relies on the minimum number of data points (min\_samples). The minimum number of points is required to consider a point as a core point and the algorithm identifies clusters based on these core points. A core point is a point that has at least the minimum number of samples (min\_samples), including itself, within its neighborhood (the region around it).

 \\ 
\hline
 
 \end{longtable}

\subsection{Time Series Similarity} \label{sec:time_series_similarity}
A time series is a sequence of observed data over time which can be defined as $ X=\{ x_1, x_2, x_3.....x_n\}$, where n is the number of observations and each $x_i$ is measured at timestamp $t_i$ and $t_{i \text{+}1} > t_i$. Retrieving the behavior of a specific data pattern over time can be very useful information. Due to the Internet of Things (IoT), there is an increasing need to understand signals from devices installed in hospitals, households, shopping malls, factories, offices, etc. These IoT devices treat the signals as time series and the challenge of time series similarity involves evaluating the likeness between two univariate time series $X_i$ and $X_j$ recorded for patients $P_i$ and $P_j$, respectively.

\begin{figure}[!htb]
\begin{center}
\includegraphics[height=250px,width=420px]{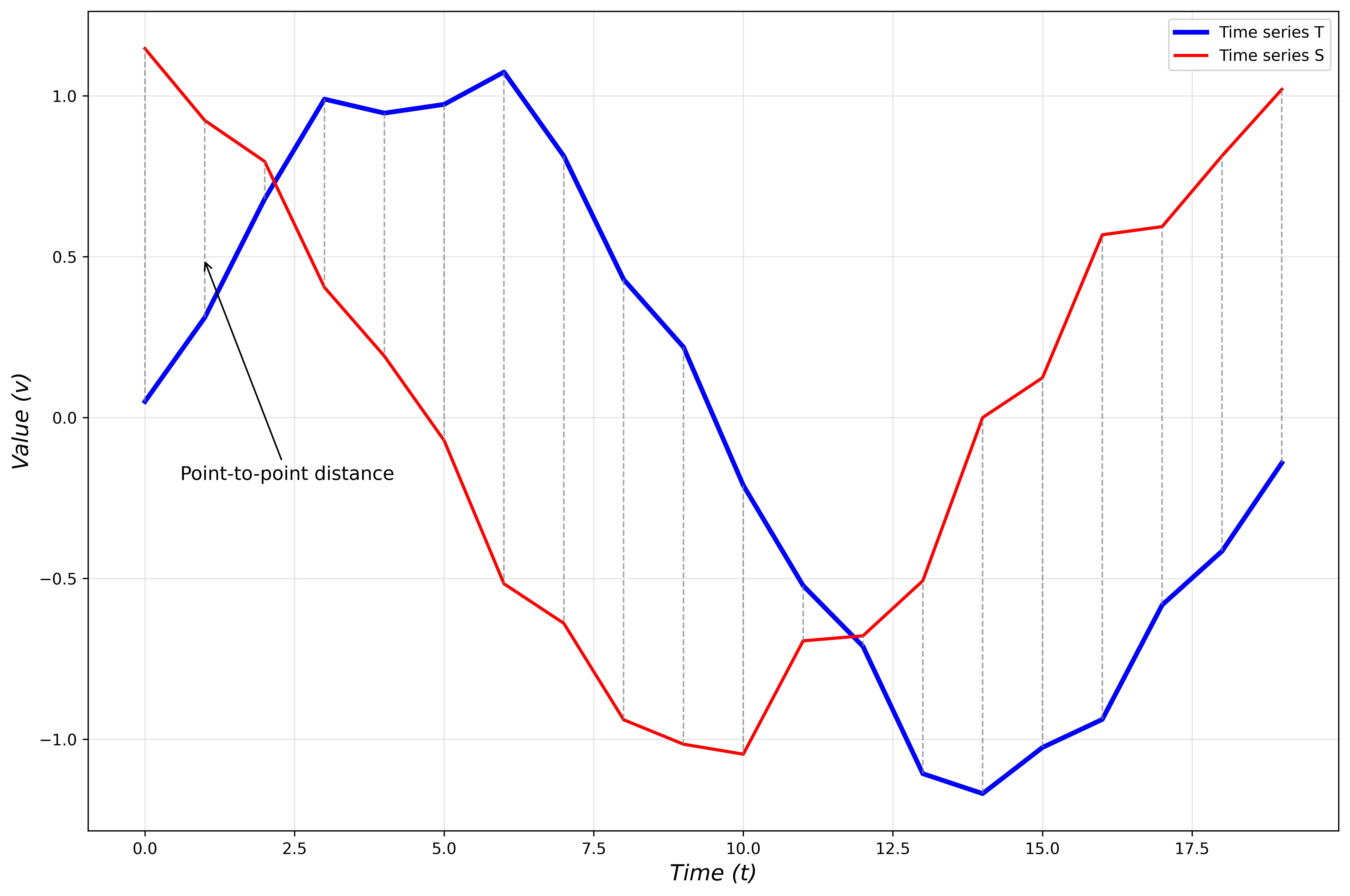}
\caption{T and S represent two time series for a specific variable v across the time axis t. The Euclidean distance is calculated as the square root of the sum of squared point-to-point distances (gray lines) between the two time series.}
\label{fig:timeseries_difference}
\end{center}
\end{figure}

The similarity or distance between two time sequences of equal length can be generated by summing the ordered distance from point to point between them (Fig. \ref{fig:timeseries_difference}). Euclidean Distance \cite{Time_Series_Matching} is the widely used distance function to calculate the similarity. However, Euclidean distance is not suitable for calculating the similarity between the two time series data. The Euclidean distance possesses several limitations: it compares only time series of equal length and does not handle outliers or noise, making it unsuitable for certain applications like identifying similar patients through time series data. To address these issues, alternative distance measurement techniques were introduced to enhance the robustness of similarity computation. In this context, we used the widely recognized Dynamic Time Warping (DTW) \cite{Keogh_Ratanamahatana_2002} to calculate the patient similarity based on the time series data. Our aim is to calculate the similarity between each pair of patients $P_i$, $P_j$  for each variate X (e.g. heart rate, respiratory rate), and produce a similarity matrix $M\big(X\big)^{N\times N}$ where N is the number of patients.

\begin{figure}[!htb]
\begin{center}
\includegraphics[height=250px,width=420px]{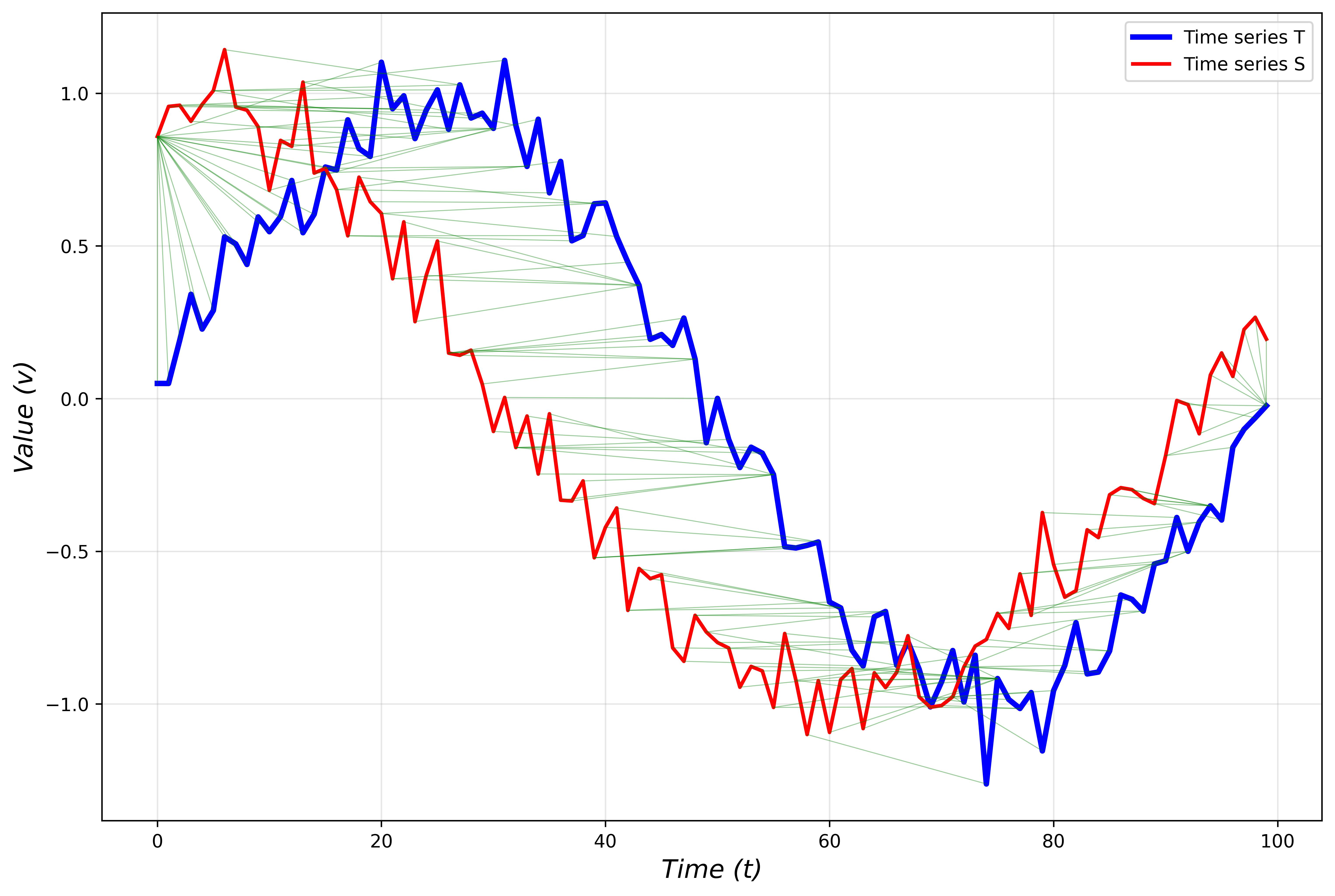}
\caption{Time series distance measurement using DTW.}
\label{fig:DTW_difference}
\end{center}
\end{figure}

\subsection{Dynamic Time Warping (DTW)} \label{sec:DTW}
DTW reduces the impact of shifts, consistent amplitude scaling, or uniform time scaling on distance comparisons. To address the challenge of differing time series lengths and overcome the drawbacks of point-to-point distance methods, researchers have recently adopted DTW \cite{Salvador2007} \cite{Keogh_Ratanamahatana_2002}, which offers increased robustness in computing time series similarity. It is also capable of comparing different lengths of time series by replacing the one-to-one point comparison utilized in Euclidean distance with a many-to-one (and vice versa) comparison. The key aspect of this distance measure is that it is able to identify similar shapes, even in the presence of transformations like shifting and/or scaling (Fig. \ref{fig:DTW_difference}) \cite{Cassisi_Montalto}. Additionally, DTW can assess the similarity between two sequences that could differ in time or speed \cite{Donald_1994}. DTW conducts temporal alignment by considering the local time shift within each series. As a result, DTW is adaptable in managing time series with varying sizes and speeds, accommodating patterns of acceleration and deceleration within the series.  
Given two time series $T = {t_1, t_2, . . . , t_n}$ and $S = {s_1, s_2, . . . , s_m}$ of length $n$ and $m$, respectively, an alignment using DTW method exploits information contained in $n\times  m$ distance matrix \cite{Cassisi_Montalto}:

\begin{equation*}
DTW_{\text{Distance-Matrix}} = 
\begin{pmatrix}
D{\left(t_1,s_1\right)} & D{\left(t_1,s_2\right)} & \cdots & D{\left(t_1,s_m\right)}  \\
D{\left(t_2,s_1\right)}  & D{\left(t_2,s_2\right)}  & \cdots & D{\left(t_2,s_m\right)}  \\
\vdots  & \vdots  & \ddots & \vdots  \\
D{\left(t_n,s_1\right)}  & D{\left(t_n,s_2\right)}  & \cdots & D{\left(t_n,s_m\right)}  
\end{pmatrix}
\end{equation*}

where Distance-Matrix(i, j) represents the distance between the $i_{th}$ point of T and the $j_{th}$ point of S $D\left(T_i, S_j\right)$, with $1 \leq i \leq n$ and $1 \leq j \leq m$. To find the minimum-distance warp path, every cell of the cost matrix must be filled.

\begin{figure}[!htb]
\begin{center}
\includegraphics[height=300px,width=300px]{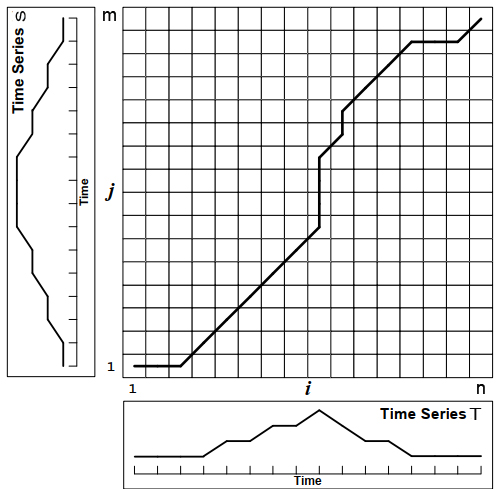}
\caption{Warping path of DTW \cite{Salvador2007}}
\label{fig:DTW_Wraping_path}
\end{center}
\end{figure}

The main aim of the DTW is to discover the warping path $W = {w_1, w_2, . . . ,w_k, . . ., w_K}$ among adjacent elements on Distance-Matrix (with $max(n, m) \le K \le m + n -1$, and $W_k = \text{Distance-Matrix}(i, j))$, aiming to minimize the subsequent function:
 
\begin{center}
 \begin{equation}
       DTW\big(T,S\big)=\min\Bigg( \sqrt{\displaystyle {\sum_{k=1}^{k} W_k}} \Bigg)  
 \end{equation}
\end{center}
  The warping path is subject to several constraints \cite{Keogh_2002} that can be efficiently computed using dynamic programming \cite{Cormen_2009}, \cite{Salvador2007}. The value at $D(i, j)$ is the minimum warp distance of two time series of lengths $i$ and $j$. A typical approach is to restrict the warping path such that it follows a direction along the diagonal (Figure \ref{fig:DTW_Wraping_path}). In this process, the value of a cell in the distance matrix can be calculated using the following equation. 

\begin{center}
 \begin{equation}
       D\big(T_i, S_j\big)=D\big(T_i, S_j\big) + \min\big[  D(T_{i-1}, S_j), D(T_i, S_{j-1}), D(T_{i-1}, S_{j-1}) \big] 
 \end{equation}
\end{center}
 
  The overall complexity of the method depends on computing all distances within the distance matrix, which is O(nm). In this case, the method takes more time for large datasets. To address this issue, we introduce distributed distance matrix computation using Spark.

\subsection{Spark Computation}\label{sec:spark}
Apache Spark is an open-source, distributed processing framework and programming model commonly used for big data workloads. It leverages in-memory caching and optimized execution for faster processing, supporting general batch processing, streaming analytics, machine learning, graph analytics, ad-hoc queries, and more. It has become one of the most popular tools for running analytics jobs and mainstream solution for big data analytics \cite{spark_1}\cite{spark_2}. In a batch-processing environment utilizing Spark, input data is stored in cost-effective storage solutions. The divide and conquer ability makes the Spark so powerful. Numerous worker nodes are created to distribute the analytics computation among them. The input data gets stored in HDFS files, then divided and distributed across the workers. After the analytic computation, the resultant data can be saved back to storage. The most time-consuming part of this research is computing DTW matrices. Therefore, to reduce the execution time we decided to adopt Spark framework for our research. %Figure \ref{fig:spark_DTW} illustrates the high lever computation in spark environment in term of DTW matrix computation. 

%\begin{figure}[!htb]
%\begin{center}
%\includegraphics[height=175px,width=400px]{images/spark_DTW.jpg}
%\caption{Spark computation}
%\label{fig:spark_DTW}
%\end{center}
%\end{figure}

\subsection{Similarity Calculation} \label{sec:simlarity_calculation}
The DTW method calculates the similarity between the target patient and all other patients within the same cluster for each univariate time series. Each distinct univariate series results in a separate distance matrix. Since we have $K$ ($K$ = 18) different univariate series, $K$ unique DTW similarity matrices have been generated. The collection of these $K$ matrices can be represented into a single matrix. Each of the $K$ matrices is calculated based on the cluster of the target patient. More formally, let $M_k$ be the $N \times T$ dimensional distance matrix computed by the DTW using univariate time series $X_k$, where N is the number of patients in the cluster, T is the number of targeted patients and $K$ is the number of variates. By combining those distance matrices, we can make a single matrix. In essence, we can utilize a fusion function $M_\mathcal{F}$:

 \begin{center}
 \begin{equation}
       M_{\mathcal{F}}=\mathcal{F} \bigg( M_1, M_2,.....,M_k \bigg)
 \end{equation}
\end{center}

Thus, the $K$ different distance matrices are consolidated into a single matrix $M_{{\mathcal{F}}}$. From this combined matrix and with the help of clustering information, we generate a similarity fusion. This similarity fusion is utilized to identify the  $\lambda$-nearest neighbors ($\lambda$=1) of each target patient.

\subsection{Neighborhood Similarity Fusion}\label{sec:Neighborhood}
For a particular target patient, $\lambda$-nearest neighborhood has been populated from each matrix. The nearest similar patient, based on the smallest distance ( we considered $\lambda$=1), was selected from the same cluster of the target patient. More formally, let $N(p,j,M)$ represent the $j$-th nearest neighbor of the patient p according to the similarity matrix $M$. Consequently, in the fusion matrix $M_\mathcal{F}$, the $\lambda$-nearest neighborhood of patient p, denoted as $N_{\lambda} (p,M)$, constitutes the set of nearest neighbors \cite{Mehedy_Masud_2020}.

 \begin{center}
 \begin{equation}
       N_{\lambda} \bigg(p,M \bigg)= \cup_{j=1}^{\lambda} N\bigg(p,j,M\bigg)
 \end{equation}
\end{center}

From the $K$ time series variates, we get $K$ sets of nearest neighbors. Together with these $K$ sets of nearest neighbors similar patients, we can make a neighborhood similarity fusion. The union of $K$ similar patient can be expressed into $S_{\mathcal{P}}$:

\begin{center}
 \begin{equation}
       S_{\mathcal{P}}=\cup_{j=1}^k P
 \end{equation}
\end{center}

\subsection{Distributed Patient Similarity Model} \label{sec:DPS}

\begin{figure}[!htb]
\begin{center}
\includegraphics[height=275px,width=450px]{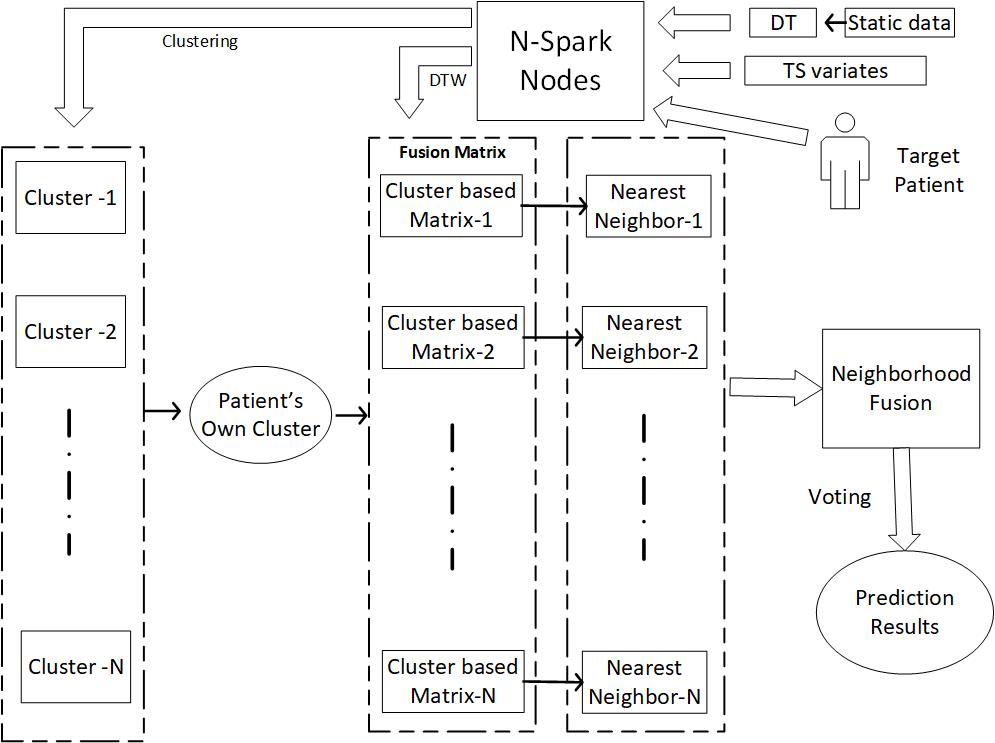}
\caption{Distributed patient similarity model}
\label{fig:Distributed_PSM}
\end{center}
\end{figure}

Figure \ref{fig:Distributed_PSM} illustrates the overall computation flow for the distributed patient similarity model. Firstly, Using the clustering algorithms, patients are partitioned into groups. Before using the static data into the clustering algorithms, DT techniques have been applied on the data. Secondly, DTW based distance matrices are computed between the targeted patient and those patients who belong to the same group of the targeted patient. Distance matrices are calculated using time series data. For each variate, a separate distance matrix will be generated. The combination of distance matrices is called fusion matrix. From each matrix we get $\lambda$ nearest neighbor patients for the targeted patient, thus the neighborhood fusion is a combination of $K$ number of nearest neighbors ($K$ is the number of variates). The class label of the majority nearest neighbors is marked as the class label of the targeted patient. The model was trained on 80\% of the dataset and tested on the remaining 20\% and its prediction performance was assessed using six evaluation metrics, which are discussed in Section \ref{subsec:evaluation_measure}..

%Firstly, DTW based distance matrices are computed using divide and conquer method on several spark nodes. Distance matrices are calculated using time series datasets. For each variate a separate distance matrix will be generated. The combination of distance matrices are called fusion matrix. Secondly, Using the clustering algorithms, patients are partitioned into groups. For each targeted patient nearest neighbor is computed from his won group. From each matrix we get one nearest neighbor patient for the targeted patient, thus the neighborhood fusion is a combination of $K$ number of nearest neighbor ($K$ is the number of variates). The class lebel of the majority nearest neighbors marked as the class lebel of the targeted patient. The clustering and prediction are carried out on the entire datasets and the prediction is based on the evaluation matrices discussed in the section \ref{subsec:evaluation_measure}.    

\subsection{Baseline Classifier}
In this research, we worked on Coronary Artery Disease and Congestive Heart Failure prediction problems and those two prediction problems are not available enough in the literature. Therefore, to compare our proposed methodology, we built a baseline method based on the methodology proposed in the study \cite{Evan2022}. Following this research, we developed several baseline models, including Naïve Bayes (NB), Logistic Regression (LR), Random forest (RF), Decision tree (DTree), Gradient boosting (GB), Extremely Randomized Trees (ExtraTrees), Adaptive Boosting (AdaBoost),  Bootstrap Aggregation (Bagging), Extreme Gradient Boosting (XGB) and  Feed-Forward Networks (FNN). To train the baseline models, both the static data and dynamic data have been utilized. We extracted the mean, standard deviation (stan. dev.), median, skewness, and median absolute deviation (MAD) from each time series variant, as outlined in \cite{Evan2022}. We used 18 time series variants, resulting in 90 features extracted from the time series data. These 90 features combined with the static data have been utilized for the baseline models mentioned above. Since the Transformer model is a powerful tool for time series analysis and long short-term memory networks (LSTM) are widely used for time series prediction, we also developed Transformer and LSTM models as baseline models using both time series and static data. The Transformer model used in this research is a simplified custom version inspired by the original Transformer architecture introduced in the paper 'Attention Is All You Need' by Vaswani et al. \cite{Vaswani_2017}. In this process, we replicated the static data (e.g., age, gender, etc.) as part of the feature vector for each time step. %For example, Time step 1: [age=25], Time step 2: [age=25], Time step 3: [age=25] for a particular patient.

\subsection{Evaluation Measure}  \label{subsec:evaluation_measure}
For measuring the performance of the models, we used the following six widely adopted evaluation metrics.

\textbf{ AUC:}  AUC refers to the Area Under the Receiver Operating Characteristic (ROC) curve, which represents the trade-off between the true positive rate and the false positive rate. AUC is not represented by a specific equation; rather, it is calculated by numerically integrating the ROC curve
   
\textbf{ Accuracy:} 
It is the accurately predicted rate of the model. Mathematically Accuracy can be expressed as:
 \begin{equation} \label{eq:Accuracy}
       Accuracy = \frac{TP+TN}{TP+FN+FP+TN}
  \end{equation} 

\textbf{Specificity:}
It is the true negative rate that is the prediction of negative records. The mathematical equation of specificity is:
 \begin{equation} \label{eq:Specificity}
       Specificity = \frac{TN}{FP+TN}
  \end{equation}
\textbf{ Precision:} It is the positive predictive value of the model. Precision can be defined as:
   \begin{equation} \label{eq:precision}
       Precision = \frac{TP}{TP+FP}
  \end{equation}
\textbf{ Recall:} Utilizing recall as an evaluation metric is appropriate when the objective is to capture the true positive rate. The equation for recall is:
  \begin{equation} \label{eq:recall}
      Recall = \frac{TP}{FN+TP}
  \end{equation}

\textbf{ F-Measure:} The F-measure is the harmonic mean of precision and recall. It becomes crucial when there is a need to balance precision and recall. An ideal model achieves an F-measure of 1. The mathematical formula for the F-measure is defined below. 

  \begin{equation} \label{eq:f-measure}
       \text{F-measure} =\frac{(2*precision*recall)}{(precision+recall)}
  \end{equation}

\subsection{Statistical Test} \label{sec:Statistical_test}
In this study, we used Friedman statistical test \cite{Janez_2006} \cite{joydebSana2022} to examine whether our proposed techniques give statistically significant different prediction performance or not. Friedman statistical test is a non-parametric statistical test and is widely used for multiple hypothesis testing. This test is reliable and does not depend on any particular data distribution. It is used to identify the significant difference among the several prediction performances \cite{joydebSana2022}. The statistical test will be performed in two phases. In the first phase, If the null-hypothesis ($H_0$) is rejected by the Friedman test, we can proceed for the second phase with a post-hoc test. In this study, we performed the Holm test for the post hoc analysis. Holm's procedure is more powerful and does not need any additional assumptions about the hypothesis testing \cite{Janez_2006}. Here, the null-hypothesis ($H_0$) is "there is no significant difference among the performance of the DT based and non-DT based methods". The tests were carried out with a significance level of $\alpha = 0.05$.

\subsection{Spark Setup and Experimental Environment}  \label{subsec:Spark_setup} 
Taking advantage of Spark's divide-and-conquer capability, we configured 3 types of clusters in our university's computational service, such as 1 master:1 cluster node, 1 master:2 cluster nodes, 1 master:3 cluster nodes. Each master and cluster node is equipped with an Intel Xeon(R) Silver 4214R CPU, having 8 cores and 16GB RAM. We conducted our experiments using Spark version 3.1.2 on Hadoop 3.2. The experiments were carried out using Python, specifically the Python version of Spark (PySpark). The PySpark console was used for execution. The code was written in Python 3.7, and Jupyter Notebook was utilized for coding. The complete set of code can be accessed through the following link: \\ {https://github.com/joysana1/PSC}.

\section{Experimental Results} \label{sec:experimental_result}
In order to measure the prediction performance, we conducted several experiments and the aims of the experiments were to predict two target features: Coronary Artery Disease, and  Congestive Heart Failure. Our proposed method is an effective fusion of time series DTW similarity and static data based clustering similarity. Distributed Spark nodes have been used for the proposed patient similarity computation technique. We used two data transformation methods and four clustering algorithms to cluster the patients. 
%Table \ref{table:DPS_F1_score} describes the overall prediction results of our proposed models. The distributed similarity model based on aWOE-\textit{K}-Means exhibits the best prediction performance for Diagnosis prediction, while the Z-score-\textit{K}-Means based distributed similarity model demonstrates better results for predicting Congestive heart failure.

\begin{table}[]
\caption{Prediction performance of distributed patient similarity models, the best result is shown in \textbf{bold-face}.}
\label{table:DPS_F1_score}
%\begin{tabular}{cccccccc}
{
\renewcommand{\arraystretch}{1.6}

\begin{tabular}{p{1.5cm}  p{1.5cm}  p{2.2cm}p{1.1cm}p{1.1cm}p{1.3cm}p{1.3cm}p{1.1cm}p{1.6cm}}

\hline\\
Target Variable & Data Transformation Method & Clustering Algorithm & AUC & Accuracy& Specificity& Precision& Recall& F-measure   \\
\hline\\
\multirow{12}{*}{CAD} &  aWOE & Spectral&0.851 &0.864 & 0.792& 0.873&0.910& 0.891      \\
                           & aWOE & \textit{K}-Means & \textbf{0.858} & \textbf{0.870}& 0.802 &\textbf{0.879} & \textbf{0.913}&   \textbf{0.896}      \\
                           & aWOE & Agglomerative      &0.849&0.860& 0.802& 0.877&0.896&   0.886      \\
                           & aWOE & OPTICS         &0.835& 0.841 & \textbf{0.811}& 0.877&0.860& 0.868      \\
                           
                           & Z-score & Spectral      &0.843 &0.857 & 0.783& 0.867&0.904& 0.885      \\
                           & Z-score & \textit{k}-Means    &0.848&0.861&  0.791& 0.872&0.906&0.889      \\
                           & Z-score & Agglomerative        &0.845&0.861 & 0.771& 0.863&0.919& 0.890      \\
                           & Z-score & OPTICS            &0.826&0.828 & 0.814&0.876 &0.838&0.856      \\
                           
                           & No-DT & Spectral      &0.621 & 0.645& 0.513 & 0.702& 0.730& 0.716     \\
                           & No-DT & \textit{k}-Means    & 0.605& 0.632 & 0.485 & 0.689 & 0.725 & 0.706      \\
                           & No-DT & Agglomerative        & 0.594& 0.616 & 0.495& 0.683 & 0.693 & 0.688      \\
                           & No-DT & OPTICS            & 0.598& 0.548& 0.81& 0.768 &  0.375 & 0.503      \\

\hline
 \multirow{12}{*}{CHF} & aWOE & Spectral &0.870 &0.886 & \textbf{0.909}& \textbf{0.791}& 0.830& 0.810      \\
                                          & aWOE & \textit{k}-Means     &\textbf{0.878}  &\textbf{0.887} & 0.900&  0.781& \textbf{0.855}& \textbf{0.816}      \\
                                          & aWOE & Agglomerative       &0.868  &0.872 &  0.879& 0.745&0.850& 0.797      \\
                                          & aWOE & OPTICS           &0.830 &0.863& \textbf{0.909}& 0.775&0.751&0.763      \\

                                           & Z-score & Spectral     &0.858 &0.878 &  0.906& 0.782&0.811& 0.796      \\
                                          & Z-score & \textit{k}-Means   &0.854 &0.875 &  0.905& 0.778&0.804& 0.791      \\
                                          & Z-score & Agglomerative      &0.8446 &0.871 &  0.908& 0.779&0.780& 0.780    \\
                                          & Z-score & OPTICS      & 0.776 & 0.842 & 0.935 & 0.798 & 0.617 & 0.696      \\

                                          & No-DT & Spectral      & 0.630 & 0.707 & 0.817 & 0.501& 0.443 & 0.470      \\
                                          & No-DT & \textit{k}-Means    & 0.618&  0.700 & 0.816 & 0.487 &0.420 & 0.451      \\
                                          & No-DT & Agglomerative        & 0.603 & 0.684 &0.798 &0.456 &0.408 & 0.431      \\
                                          & No-DT & OPTICS            &0.618 &0.514 & 0.369 & 0.362 & 0.867 & 0.511      \\

\hline
  
\end{tabular}
}
\end{table}

\subsection{Performance comparison} \label{sec:F1_comparison}
Table \ref{table:DPS_F1_score} represents the prediction results of our proposed models. For CAD, aWOE-K-means based PSC achieves the highest prediction performance in terms of all measurement metrics except specificity. The highest specificity is achieved by the aWOE-OPTICS based model with a value of 0.811. For Congestive Heart Failure (CHF), aWOE-K-means based model also shows the best prediction results across all evaluation metrics, with the exception of specificity. The highest specificity was achieved by aWOE-Spectral and aWOE-OPTICS based models with a value of 0.909. Considering all prediction outcomes, it is evident that aWOE based models achieve superior prediction performance.

\begin{figure}[h!]
    \centering
    \begin{subfigure}{1.0\textwidth}
        \includegraphics[width=\linewidth]{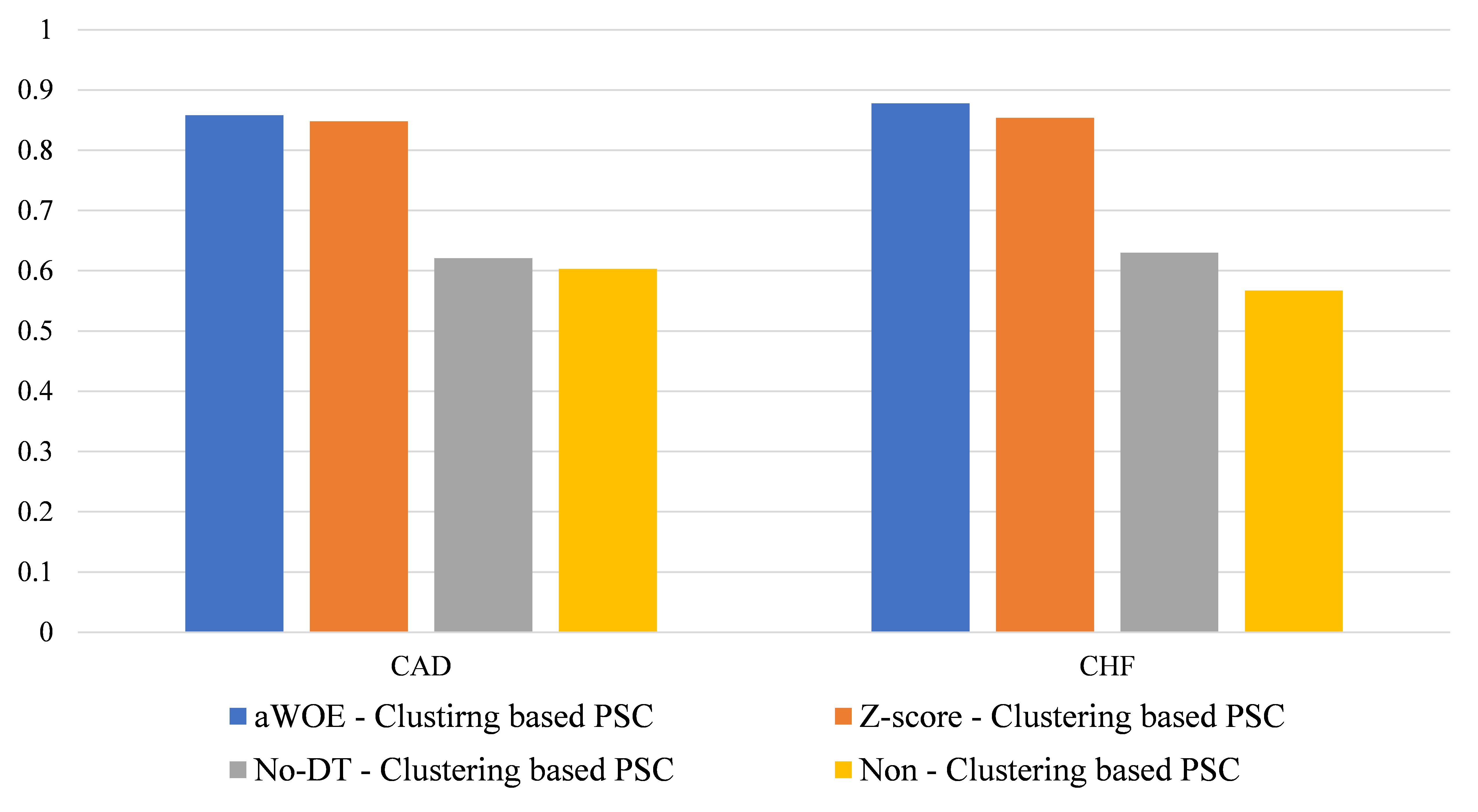}
        \caption{Comparison in terms of AUC}
    \end{subfigure}
    \begin{subfigure}{1.0\textwidth}
        \includegraphics[width=\linewidth]{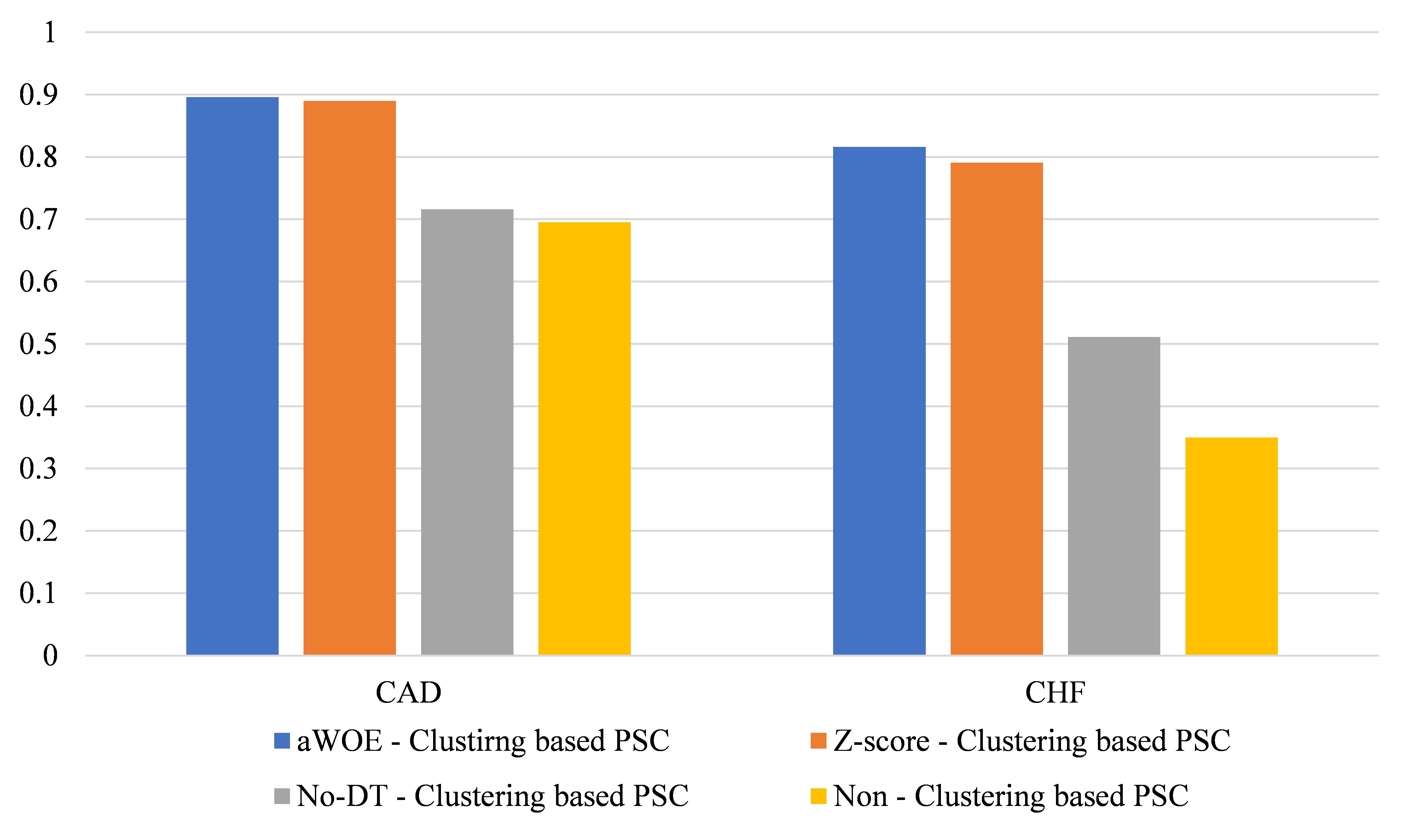}
        \caption{Comparison in terms of F-measure}
    \end{subfigure}

    \caption{Comparison between DT-clustering based, No-DT-clustering based and Non-clustering based models.}
     \label{fig:clustering-No-clustering-comparison}
\end{figure}

To evaluate the effectiveness of our proposed model, we implemented non-clustering-based and No-DT-clustering-based similarity methods and compared their performance with our proposed technique. Figure \ref{fig:clustering-No-clustering-comparison} illustrates the comparison between our proposed PSC, the No-DT-clustering-based PSC, and the non-clustering-based PSC in terms of AUC and F-measure. In all cases, the proposed methods consistently outperformed the alternatives. Specifically, the aWOE-clustering based method achieved the best performance for both CAD and CHF, with F-measure scores of 0.896 (CAD) and 0.816 (CHF), as shown in Table \ref{table:DPS_F1_score}. Regarding AUC, the aWOE-clustering based technique recorded 0.858 for CAD and 0.878 for CHF. The K-means clustering based approach achieved the highest performance, followed by Spectral Clustering. For both prediction tasks, the aWOE-based patient similarity techniques delivered the highest prediction accuracy.

% \begin{figure}[h!]
%    \centering
%    \begin{subfigure}{0.45\textwidth}
%        \includegraphics[height=225px,width=225px]{images/clustering-vs-no-clustering-Kmeans-AUC.jpg}
%        \caption{Comparison in terms of AUC}
%    \end{subfigure}
%    \begin{subfigure}{0.45\textwidth}
%        \includegraphics[height=225px,width=225px]{images/clustering-vs-no-clustering-Kmeans-F-measure.jpg}
%        \caption{Comparison in terms of F-measure}
%    \end{subfigure}

%    \caption{Comparison among clustering based models and Non-clustering based model.}
%     \label{fig:clustering-No-clustering-comparison}
%\end{figure}

% \begin{figure}[h!]
%    \centering
%    \begin{subfigure}{0.45\textwidth}
%        \includegraphics[height=225px,width=225px]{images/DT-vs-no-DT-Kmeans-AUC.jpg}
%        \caption{Comparison in terms of AUC}
%    \end{subfigure}
%    \begin{subfigure}{0.45\textwidth}
%        \includegraphics[height=225px,width=225px]{images/DT-vs-no-DT-Kmeans-F-measure.jpg}
%        \caption{Comparison in terms of F-measure}
%    \end{subfigure}
%
%    \caption{Comparison among DT based techniques and No-DT based technique.}
%     \label{fig:DT_VS_NO_DT_methods_comparison}
%\end{figure}

 \subsection{Effect of Number of Clusters in Prediction Performance} \label{sec:effect_of_cluster_size}
\begin{figure}[!htb]
\begin{center}
\includegraphics[height=225px,width=380px]{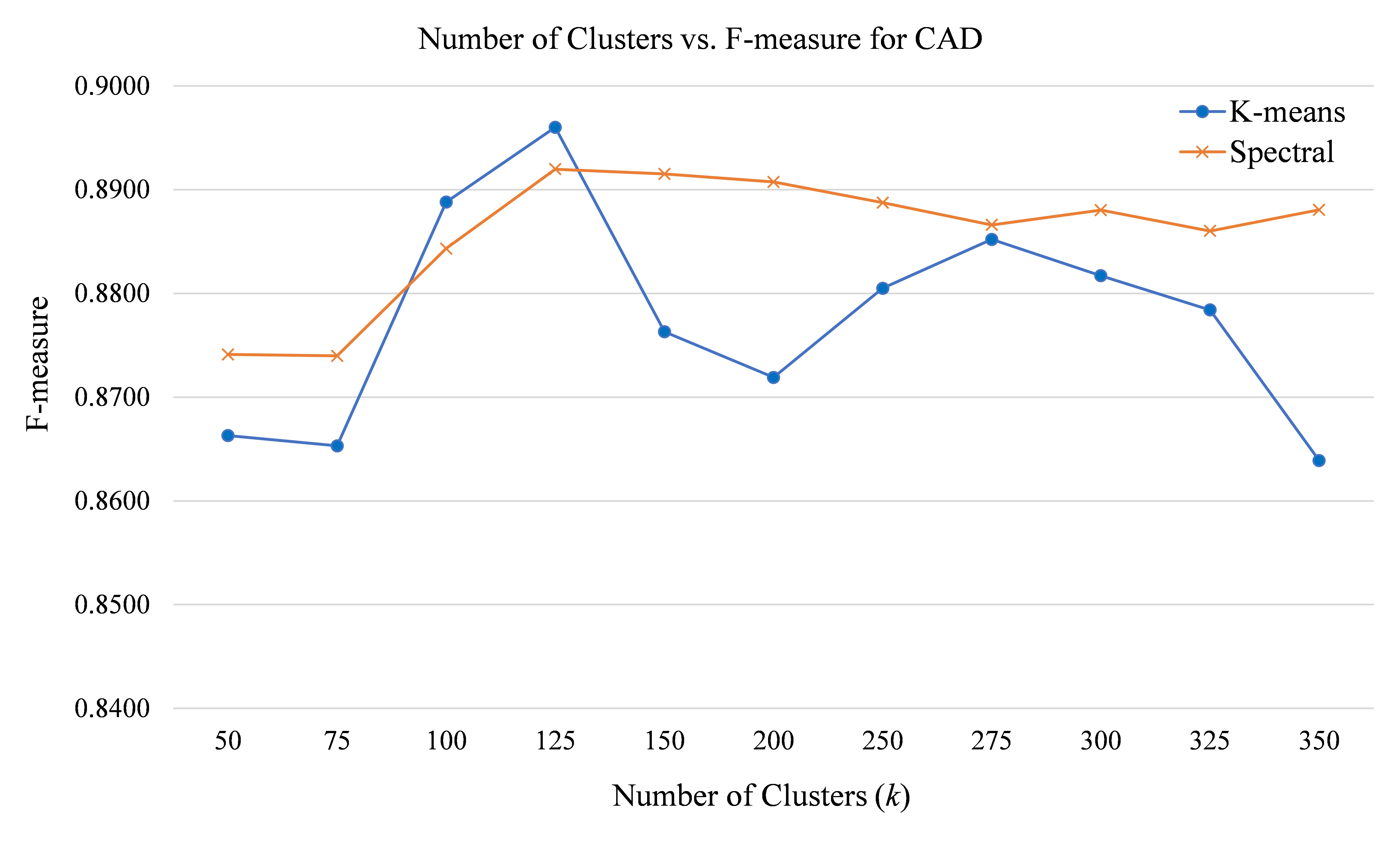}
\caption{Cluster count versus prediction performance using $K$-means and Spectral clustering for CAD.}
 \label{fig:cluster_size_vs_f-measure-aWOE-kmean}
\end{center}
\end{figure}

\begin{figure}[!htb]
\begin{center}
\includegraphics[height=225px,width=380px]{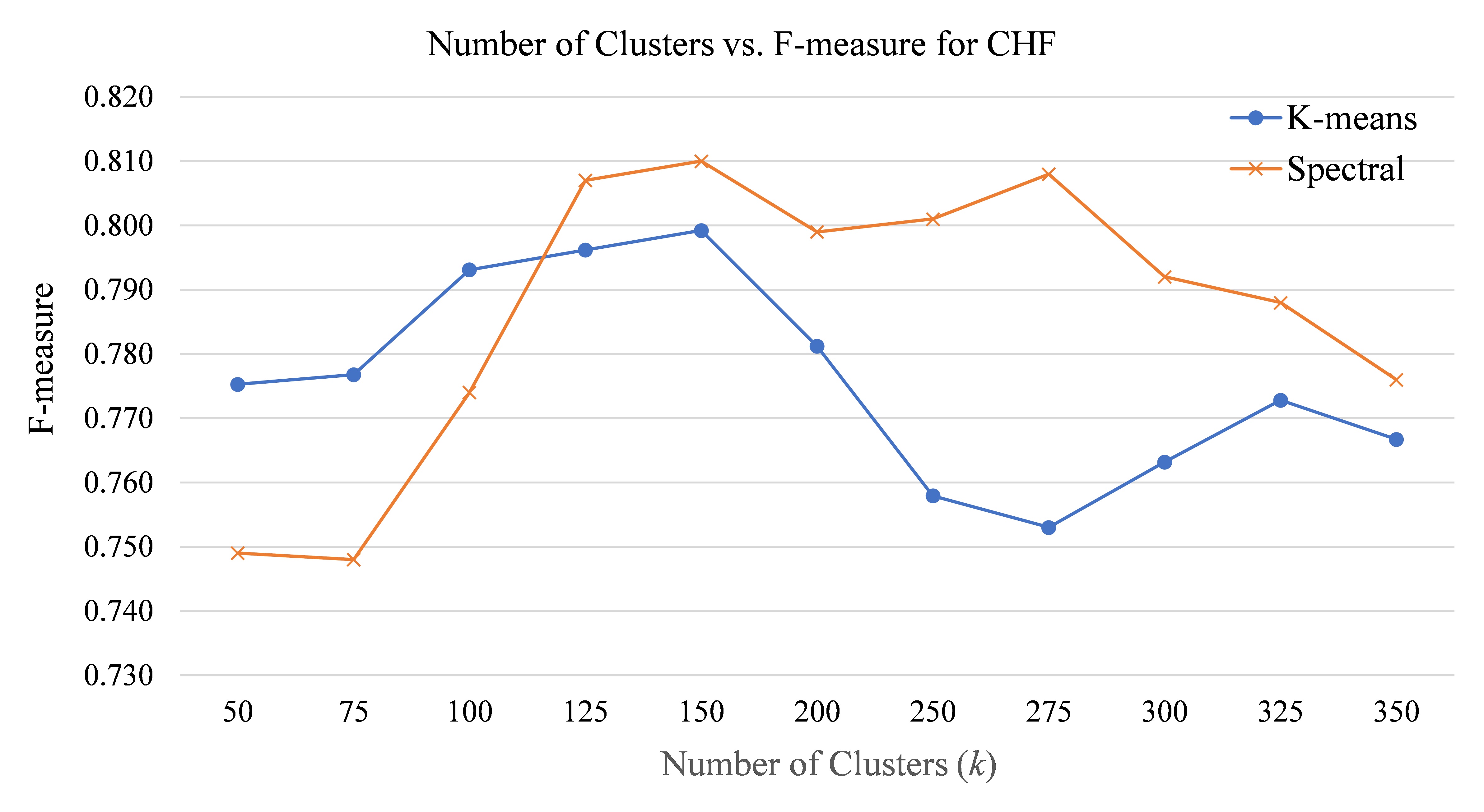}
\caption{Cluster count versus prediction performance using $K$-means and Spectral clustering for CHF.}
 \label{fig:cluster_size_vs_f-measure-aWOE-Spectral}
\end{center}
\end{figure}

 Figure \ref{fig:cluster_size_vs_f-measure-aWOE-kmean} and \ref{fig:cluster_size_vs_f-measure-aWOE-Spectral} illustrate how the prediction performance varies with the number of clusters in terms of F-measure for $K$-means and Spectral clustering in the cases of CAD and CHF, respectively. We discuss only the two clustering algorithms here, because, $K$-means clustering based method achieved the highest prediction results and the performance of Spectral clustering based method is very close to the $K$-means based method. We performed the clustering and evaluation process five times, and based on the average performance, we generated the graphs. We notice that as the cluster size ($k$) increases, the F-measure rises as well, but after reaching a certain value of $k$, the F-measure starts to decline and continues this pattern. Both K-means clustering and Spectral clustering achieved the highest prediction results when the number of clusters was 125 for Coronary Artery Disease (CAD) and 150 for Congestive Heart Failure (CHF), respectively.

 \subsection{Effect of time series data length in prediction performance} \label{sec:effect_of_time_series_data_length}
\begin{figure}[!htb]
\begin{center}
\includegraphics[height=225px,width=400px]{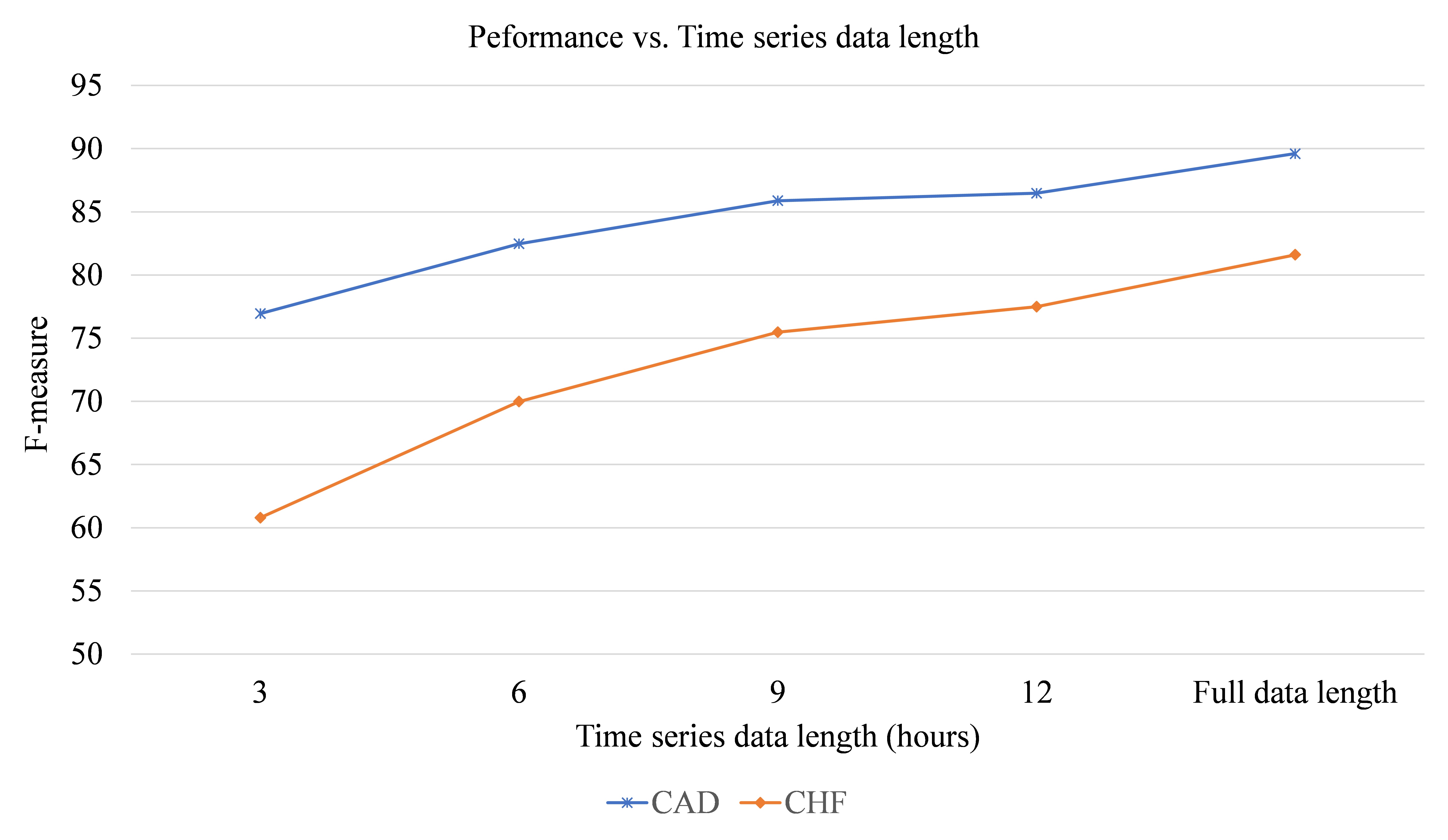}
\caption{Prediction performance utilizing various time series data lengths in terms of F-measure.}
\label{fig:time_series_data_length-vs-F-measure}
\end{center}
\end{figure}

 In this section, we evaluate the performance of our proposed patient similarity computation model by testing it with varying lengths of time series data combined with static features. Our experiments focus on observation windows of 3, 6, 9, and 12 hours to assess how the duration of time series data influences the model’s predictive performance. The results in Figure \ref{fig:time_series_data_length-vs-F-measure} demonstrate a consistent trend: as the length of the time series data increases, the model’s performance improves significantly. For the CAD, the model achieved F-measures of 76.95\%, 82.47\%, 85.87\%, and 86.47\% for 3, 6, 9, and 12 hours of observation data, respectively. Similarly, for the CHF, the model attained F-measures of 60.79\%, 69.98\%, 75.47\%, and 77.48\% for the same observation windows. These findings highlight the importance of longer time series data in capturing the temporal dynamics of these conditions, enabling the model to compute more accurate patient similarities.

Interestingly, when comparing the performance of the 12-hour observation window to the entire time series data, the differences in F-measure were relatively small. For CAD, the difference was 3.13, while for CHF, it was 4.12. This suggests that a 12-hour observation window provides a robust approximation of the full time series data, achieving near-optimal performance while reducing computational complexity and data collection requirements. It is notable that for CAD, the F-measure increased by nearly 10 percentage points from 3 to 12 hours, while for CHF, the increase was approximately 17 percentage points. Another key observation here is that the 12-hour window strikes a balance between performance and practicality, making it a viable option for real-world deployment in clinical settings where data collection and processing constraints exist. These findings have significant implications for patient similarity models in healthcare. Shorter observation windows can still deliver strong performance, enabling model deployment in environments with limited data availability, such as emergency departments.

\subsection{Execution time comparison} \label{sec:executin_time_comparison}
\begin{figure}[!htb]
\begin{center}
\includegraphics[height=225px,width=375px]{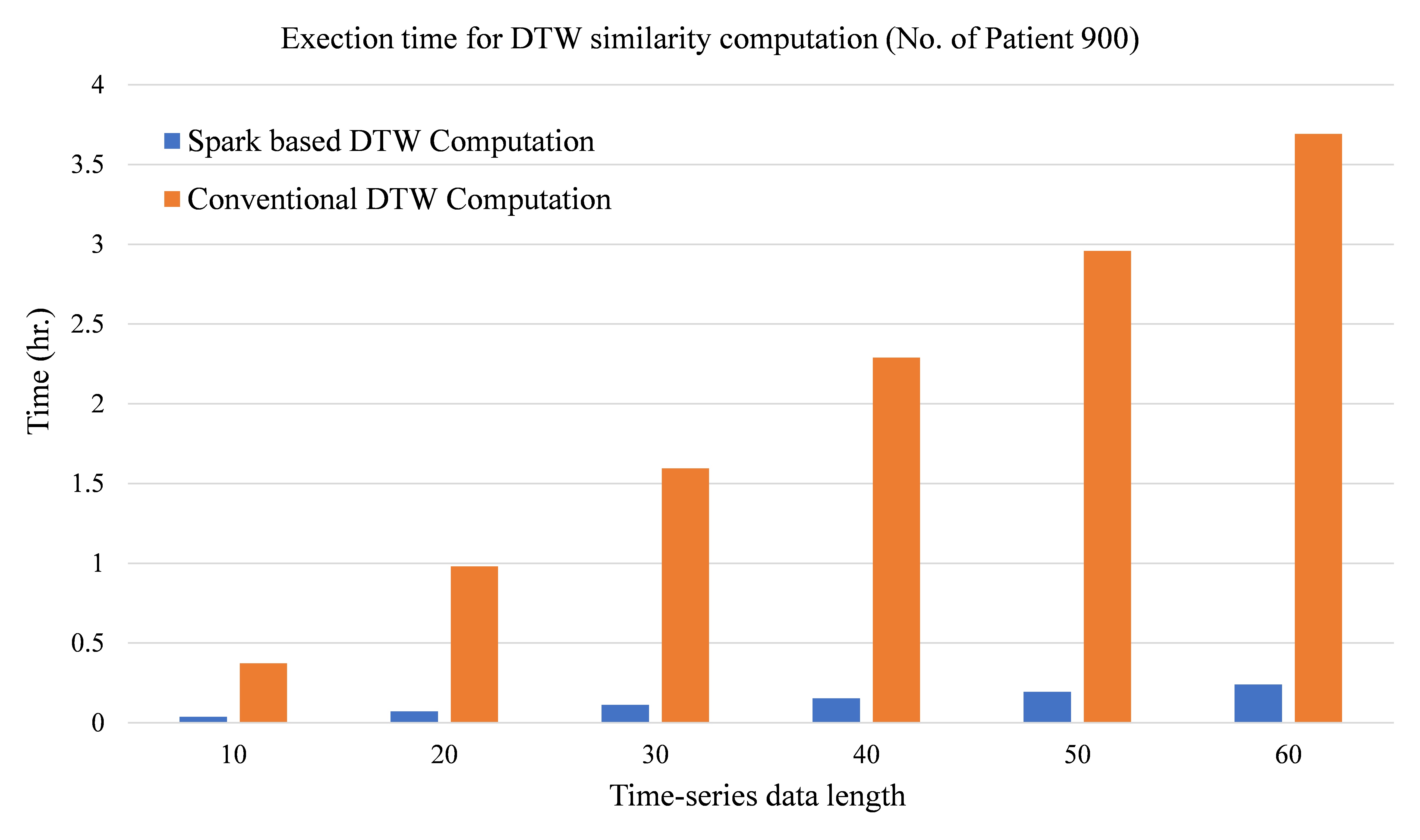}
\caption{Comparison: Execution time Vs. Time-series data length}
\label{fig:executon-time-vs-TS-length}
\end{center}
\end{figure}

\begin{figure}[!htb]
\begin{center}
\includegraphics[height=225px,width=375px]{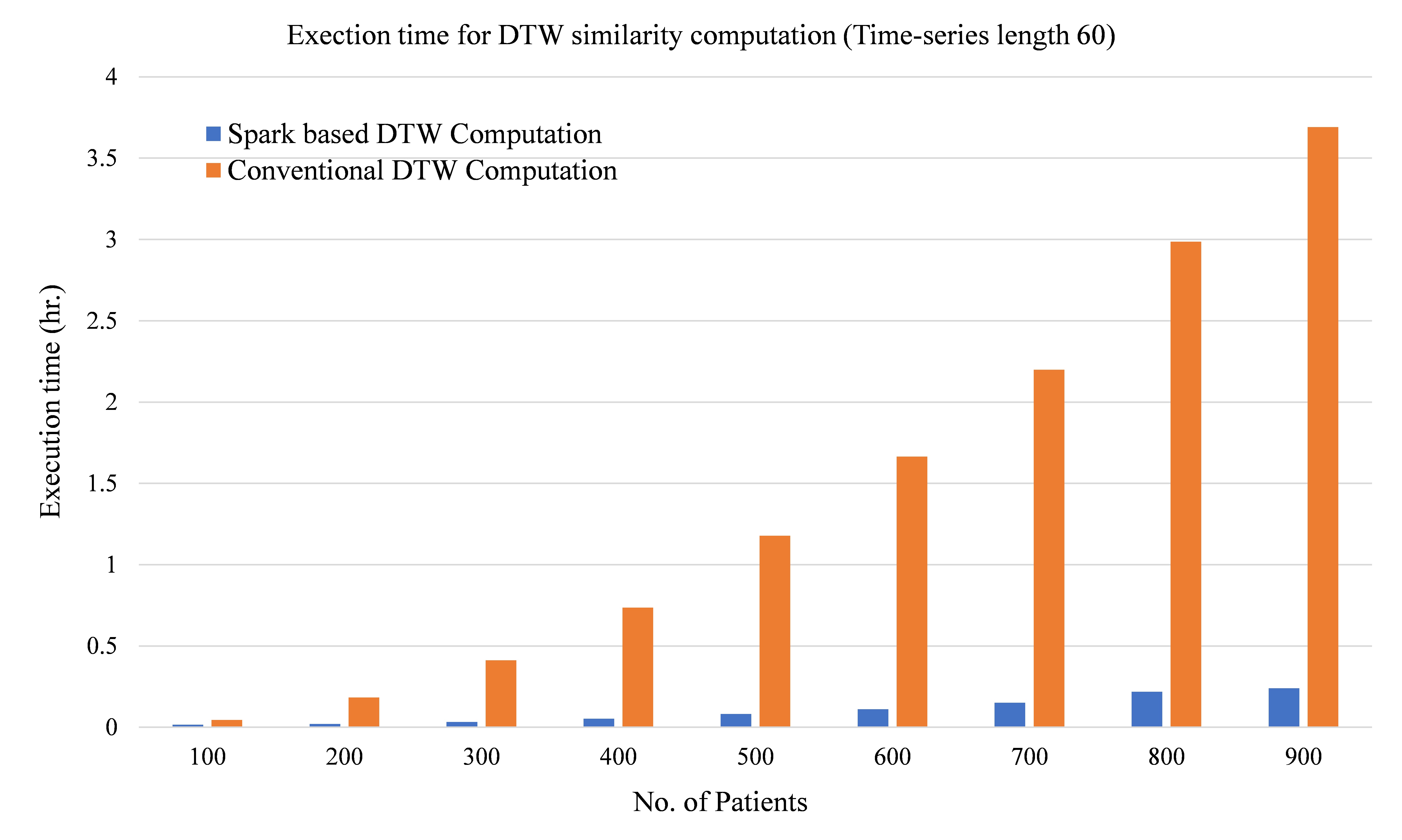}
\caption{Comparison: Execution time Vs. number of patient size}
\label{fig:executon-time-vs-patient-size}
\end{center}
\end{figure}

\begin{figure}[!htb]
\begin{center}
\includegraphics[height=225px,width=375px]{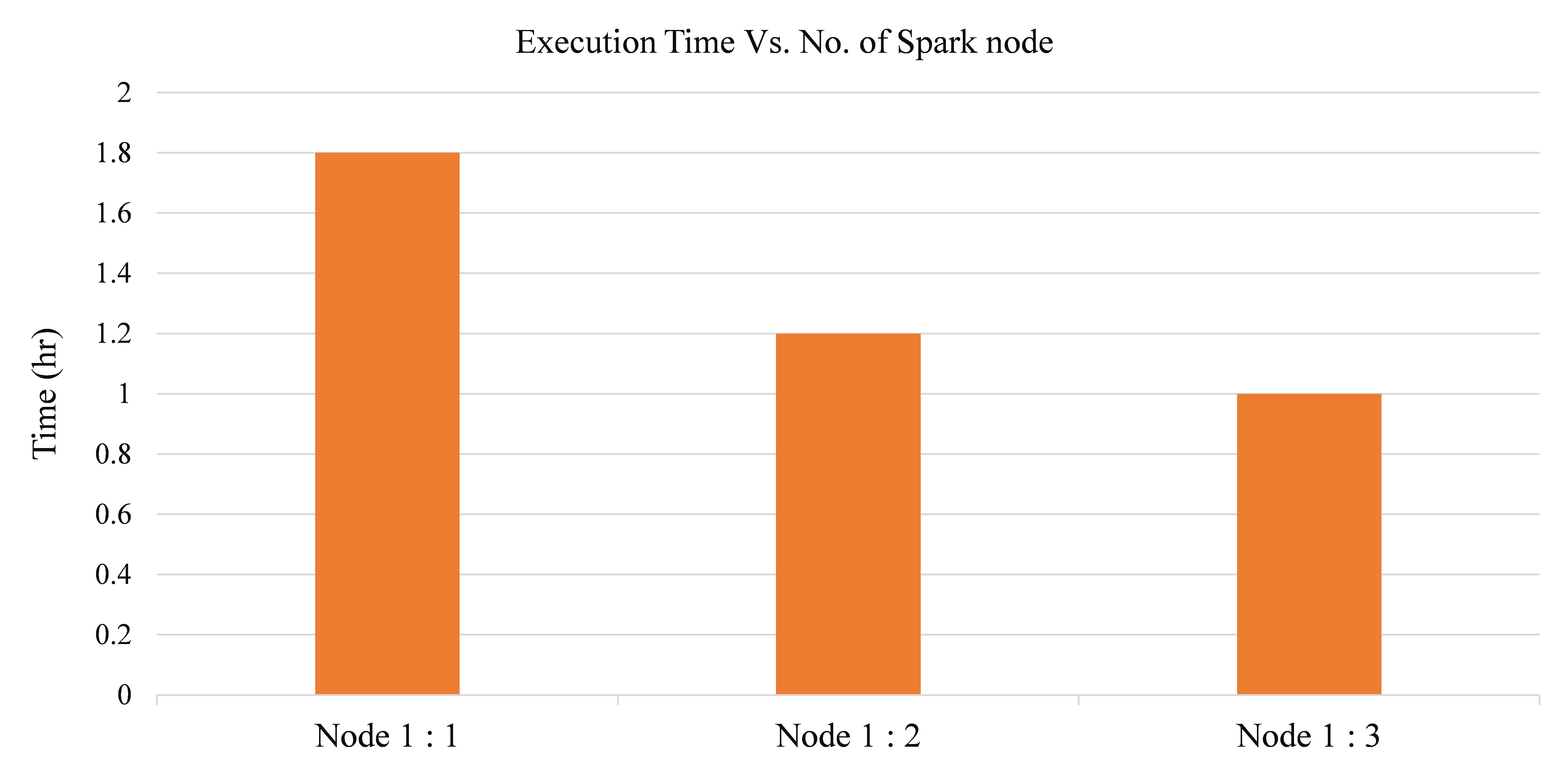}
\caption{Comparison: Execution time Vs. number of Spark nodes}
\label{fig:execution-time-spark-cluster}
\end{center}
\end{figure}

To observe, how Spark distributed computing helps to reduce the computation time for the patient similarity prediction, we performed several experiments. In big data context, conventional DTW computation may struggle to deliver real-time results. In such cases, Spark-based DTW computation can offer a solution.  Figure \ref{fig:executon-time-vs-TS-length} and \ref{fig:executon-time-vs-patient-size} demonstrate that distributed Spark-based computation reduces the execution time of DTW similarity calculation. In Figure \ref{fig:executon-time-vs-TS-length}, we altered the length of time series data while maintaining the same number of patients (900). The illustration indicates that Spark-based DTW computation requires significantly less time compared to conventional DTW computation. This observation is also evident from Figure \ref{fig:executon-time-vs-patient-size}, where we modified the number of patients while keeping the length of time series data constant. Another notable observation from Figures \ref{fig:executon-time-vs-TS-length} and \ref{fig:executon-time-vs-patient-size} is that with larger data sizes, the Spark-based DTW takes less time compared to conventional DTW computation. Our proposed similarity process has been performed in the Spark environment. The most time-intensive aspect in the similarity prediction framework is to construct the distance calculation using the DTW. So, we presented only execution time comparison of the DTW computation. Figure \ref{fig:execution-time-spark-cluster} shows another experimental result where it represents different Spark nodes size achieve different execution times. The Figure \ref{fig:execution-time-spark-cluster} illustrates that as the number of Spark nodes increases, the execution time decreases. The DPSC takes only 58 seconds for 100 patients when the time series data length is 60. Hence, it is evident that in a practical context, the prediction response time for the proposed DPSC in identifying similarity for an individual patient will be very low. From this experiment, it is observed that increasing the number of nodes in the Spark distribution system can potentially fulfill the real-time demand for patient similarity prediction and can potentially save many lives, especially among emergency patients.

\section{Statistical significance analysis} \label{sec:performance_analysis}
To assess the statistical significance of the improvements brought by our proposed techniques, we conducted the Friedman statistical test followed by the Holm post-hoc test. The Friedman test is a non-parametric statistical test and does not require any particular distribution of data. Holm's procedure is a robust post-hoc test that does not require any additional assumptions for hypothesis testing. 
%Here, the null-hypothesis ($H_0$) is there is no significant different between the performance of the proposed DT based DPSC and the non-DT based DPSC techniques. The statistical test has been performed at the significance level, $\alpha$=0.05.

The Friedman ranks of all three methods across all prediction problems and clustering algorithms are shown in the table \ref{table:ranking}. The computed Friedman test statistic is 14.25 and the P-value is 0.000804. Therefore, according to the Friedman test rule, at the significance level $\alpha =0.5$, the null-hypothesis ($H_0$) is rejected. Subsequently, the post-hoc Holm test has been performed. The table \ref{table:holm_test} represents the P-values computed by Holm test with adjusted $\alpha$. The Holm procedure rejects those hypotheses that have P-value smaller than the adjusted $\alpha$ values. The rank table \ref{table:ranking} and the post-hoc comparison table \ref{table:holm_test} illustrate that our proposed aWOE based DPSC method performs significantly better than the other methods.

\begin{table}[!htp]
\centering
\begin{tabular}{|c|c|}\hline
Method & Ranking\\\hline
aWOE based DPSC & 1.125\\
Z-score based DPSC & 1.875\\
NON-DT based DPSC & 3\\
\hline
\end{tabular}
\caption{Average Rankings of the methods}
\label{table:ranking}
\end{table}

\begin{table}[!htp]
\centering
\scriptsize
{
\renewcommand{\arraystretch}{1.5}
\begin{tabular}{cccccc}
 
$i$ & Method & $z=(R_0 - R_i)/SE$ & P-value & Holm (adjusted $\alpha$) & Hypothesis ($\alpha=0.05$)\\
\hline 
1 & aWOE based DPSC vs. NON-DT based DPSC & 3.75 & 0.000177 & 0.016667 & Rejected\\
2 & Z-score based DPSC vs. NON-DT based DPSC & 2.25 & 0.024449 & 0.025 & Rejected \\
3 & aWOE based DPSC vs. Z-score based DPSC & 1.5 & 0.133614 & 0.05 & Rejected \\

\hline
 
\end{tabular}
}
\caption{Holm test P-values table for $\alpha=0.05$}
\label{table:holm_test}
\end{table}

\section{Comparison with previous studies and baseline models} \label{sec:comparison_previous_study}
Table \ref{table:DPS_F1_score_comparision} shows the performance comparison among our proposed DT based distributed patient similarity computation models and conventional patient similarity computation technique \cite{Mehedy_Masud_2020} and the baseline models. The previous study \cite{Mehedy_Masud_2020} provided only F-measure and did not work on Congestive Heart Failure prediction. On the other hand, we worked on Coronary Artery Disease and Congestive Heart Failure prediction problems and assessed the performance of the models using six evaluation metrics ( Table \ref{table:DPS_F1_score}). We prepared several baseline models (NB, LR, RF, DTree, GB, XGB, FNN, ExtraTrees, Bagging (RF), AdaBoost, and FNN) for our prediction problems following the research work \cite{Evan2022}. Since we have time series data, we also developed LSTM and Transformer models as baseline models. From the comparison table \ref{table:DPS_F1_score_comparision}, we found that our proposed models show much higher results than the conventional patient similarity computation model \cite{Mehedy_Masud_2020} and all baseline models for both Coronary Artery Disease and Congestive Heart Failure in terms of AUC, accuracy, and F-measure. For Coronary Artery Disease, our proposed DPSC demonstrates an observable improvement of up to 12.6\% in terms of F-measure compared to the previous study \cite{Mehedy_Masud_2020}. Compared to the baseline models, the proposed model improves performance by up to 10.1\%, 9.1\%, and 7.0\% in terms of AUC, accuracy, and F-measure, respectively. In the case of Congestive Heart Failure, our proposed method achieves performance enhancements of up to 13.9\%, 9.1\%, and 3.0\% in terms of AUC, accuracy, and F-measure, respectively.

% \usepackage{multirow}
\begin{table}[]
\begin{center}
\caption{Comparison among the DT based distributed patient similarity models, conventional patient similarity model \cite{Mehedy_Masud_2020} and baseline models, the best result is shown in \textbf{bold-face}.}
\label{table:DPS_F1_score_comparision}
{
\renewcommand{\arraystretch}{1.5}
\begin{tabular}{|l|lll|lll|}
\hline
\multirow{2}{*}{Ref Study}                     & \multicolumn{3}{l|}{Coronary Artery Disease}                                       & \multicolumn{3}{l|}{Congestive Heart Failure}                        \\ \cline{2-7} 
                                               & \multicolumn{1}{l|}{AUC} & \multicolumn{1}{l|}{Accuracy} & F-Measure & \multicolumn{1}{l|}{AUC} & \multicolumn{1}{l|}{Accuracy} & F-Measure \\ \hline
aWOE based DPSC (Proposed model)    & \multicolumn{1}{l|}{\textbf{0.858}}   & \multicolumn{1}{l|}{\textbf{0.870}}        &     \textbf{0.896}      & \multicolumn{1}{l|}{\textbf{0.878}}   & \multicolumn{1}{l|}{\textbf{0.887} }         & \textbf{0.816}        \\ \hline
Z-score based DPSC (Proposed model) & \multicolumn{1}{l|}{0.848}   & \multicolumn{1}{l|}{0.861}        &      0.889     & \multicolumn{1}{l|}{0.858}   & \multicolumn{1}{l|}{0.878}         & 0.796          \\ \hline

N-Pop PSC \cite{Mehedy_Masud_2020}             & \multicolumn{1}{l|}{--}   & \multicolumn{1}{l|}{--}        &      0.77       & \multicolumn{1}{l|}{--}   & \multicolumn{1}{l|}{--}         &    --    \\ \hline

NB          & \multicolumn{1}{l|}{0.582}   & \multicolumn{1}{l|}{0.523}        &    0.446       & \multicolumn{1}{l|}{0.594}   & \multicolumn{1}{l|}{ 0.481}         & 0.480         \\ \hline

LR          & \multicolumn{1}{l|}{0.689}   & \multicolumn{1}{l|}{0.726}        &    0.792       & \multicolumn{1}{l|}{0.674}   & \multicolumn{1}{l|}{0.779}         & 0.525         \\ \hline

RF            & \multicolumn{1}{l|}{0.739}   & \multicolumn{1}{l|}{0.764}        &    0.815       & \multicolumn{1}{l|}{0.739}   & \multicolumn{1}{l|}{0.796}         & 0.628         \\ \hline

DTree            & \multicolumn{1}{l|}{0.696}   & \multicolumn{1}{l|}{0.711}        &   0.764       & \multicolumn{1}{l|}{0.662}   & \multicolumn{1}{l|}{0.713}         & 0.518         \\ \hline

GB            & \multicolumn{1}{l|}{0.741}   & \multicolumn{1}{l|}{0.769}        &    0.820       & \multicolumn{1}{l|}{0.717}   & \multicolumn{1}{l|}{0.794}         & 0.597         \\ \hline

XGB             & \multicolumn{1}{l|}{0.741}   & \multicolumn{1}{l|}{0.762}        &    0.811       & \multicolumn{1}{l|}{0.710}   & \multicolumn{1}{l|}{0.768}         & 0.584         \\ \hline

FNN               & \multicolumn{1}{l|}{0.691}   & \multicolumn{1}{l|}{0.738}        &    0.808    & \multicolumn{1}{l|}{0.50}   & \multicolumn{1}{l|}{0.718}         & 0.600         \\ \hline

ExtraTrees            & \multicolumn{1}{l|}{0.744}   & \multicolumn{1}{l|}{0.768}        &    0.818       & \multicolumn{1}{l|}{0.719}   & \multicolumn{1}{l|}{0.782}         & 0.597         \\ \hline

Bagging (RF)              & \multicolumn{1}{l|}{0.757}   & \multicolumn{1}{l|}{0.779}        &     0.826      & \multicolumn{1}{l|}{0.727}   & \multicolumn{1}{l|}{0.788}         & 0.61          \\ \hline

AdaBoost              & \multicolumn{1}{l|}{0.748}   & \multicolumn{1}{l|}{0.768}        &      0.815     & \multicolumn{1}{l|}{0.722}   & \multicolumn{1}{l|}{0.786}         & 0.602         \\ \hline

LSTM              & \multicolumn{1}{l|}{0.737}   & \multicolumn{1}{l|}{0.76}        &    0.81       & \multicolumn{1}{l|}{0.732}   & \multicolumn{1}{l|}{0.76}         & 0.60         \\ \hline

Transformer              & \multicolumn{1}{l|}{0.731}   & \multicolumn{1}{l|}{0.76}        &    0.812       & \multicolumn{1}{l|}{0.731}   & \multicolumn{1}{l|}{0.762}         & 0.813        \\ \hline

\end{tabular}
}
\end{center}
\end{table}

\section{Impact of adaptive Weight-of-Evidence on model output} \label{sec:impact_aWOE}
\begin{figure}[h!]
    \centering
    \begin{subfigure}{0.48\textwidth}
        \includegraphics[height=150px,width=200px]{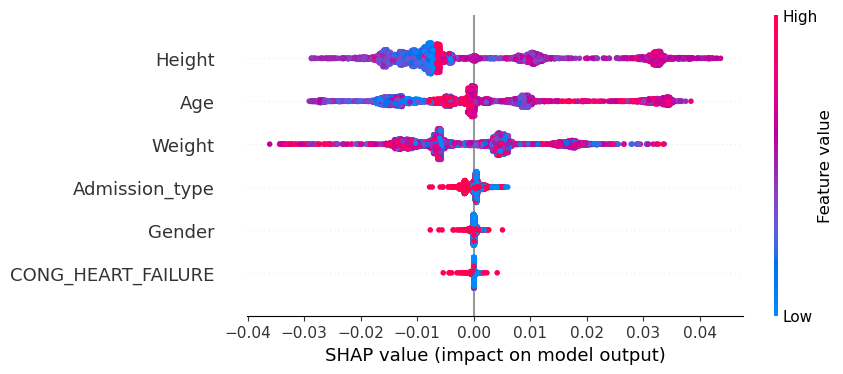}
         \caption{RAW (NO-DT) based Clustering}
    \end{subfigure}
    \begin{subfigure}{0.48\textwidth}
        \includegraphics[height=150px,width=200px]{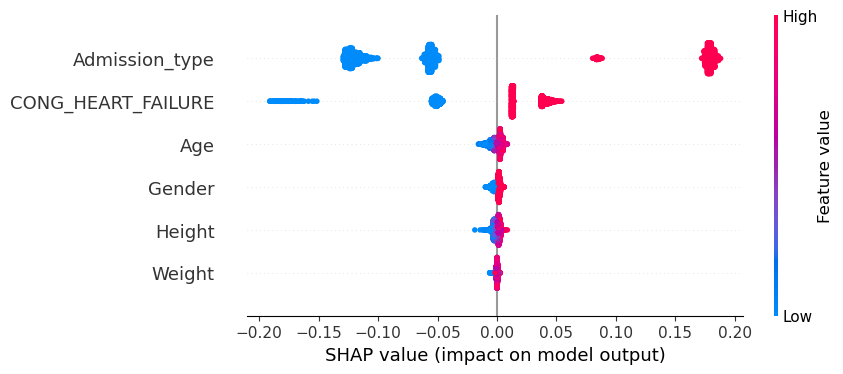}
        \caption{aWOE based Clustering}
    \end{subfigure}

    \caption{The Beeswarm SHAP plot using $K$-means Clustering in terms of Coronary Artery Disease.}
     \label{fig:shap_diagonosis}
\end{figure}

\begin{figure}[h!]
    \centering
    \begin{subfigure}{0.48\textwidth}
        \includegraphics[height=150px,width=200px]{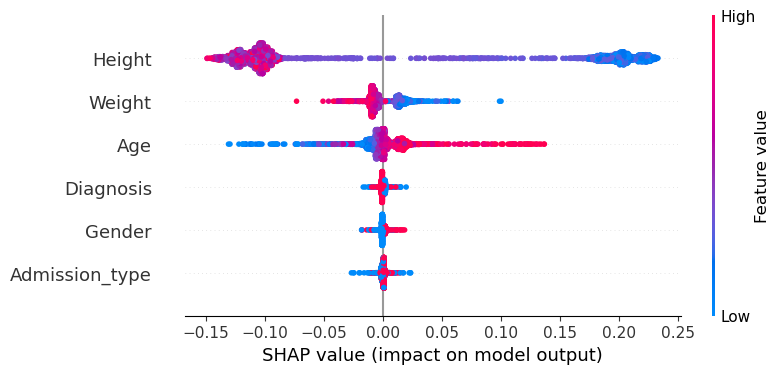}
         \caption{RAW (NO-DT) based Clustering}
    \end{subfigure}
    \begin{subfigure}{0.48\textwidth}
        \includegraphics[height=150px,width=200px]{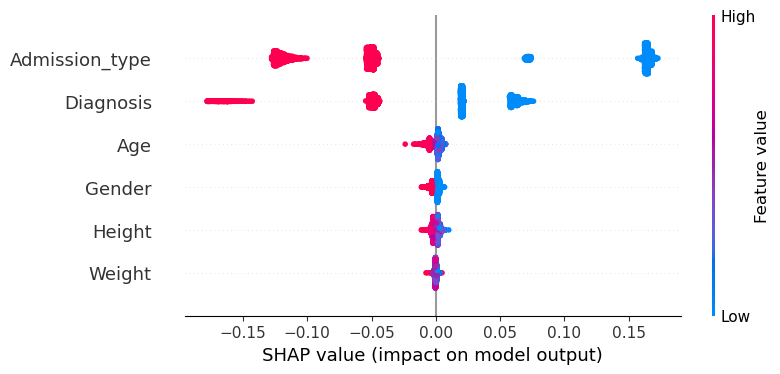}
        \caption{aWOE based Clustering}
    \end{subfigure}

    \caption{The Beeswarm SHAP plot using $K$-means Clustering in terms of Congestive Heart Failure.}
     \label{fig:shap_Congestive}
\end{figure}

 Data transformation methods have been applied on static data before feeding the data to the clustering algorithms. To show the contribution or the importance of each feature to the prediction of the clusters, SHAP (Shapley Additive eXplanation) analysis has been implemented. Figures \ref{fig:shap_diagonosis} and \ref{fig:shap_Congestive} represent the Beeswarm SHAP plots for both the RAW and aWOE, using $k$-means clustering, with respect to Coronary Artery Disease and Congestive Heart Failure, respectively. Here, the features are ordered from the highest to the lowest effect on the prediction. From the SHAP plots, we observed that Admission type is the most important feature for the aWOE based Coronary Artery Disease prediction. For the Congestive Heart Failure prediction, Coronary Artery Disease is the most contributing feature followed by Admission type. From the SHAP analysis, it is remarkable that for both cases, patient background or medical records contribute more than the patient demographic information for the prediction when aWOE is applied. It is well-known that current medical records provide detailed and actionable insights into a patient’s health status. Medical data such as admission, diagnoses, and other factors are directly linked to patient outcomes and are critical for making accurate predictions. While demographic features like height, weight, age, and gender offer a broader context, they do not provide the level of specificity needed to understand individual health trajectories. SHAP analysis confirms that in aWOE based model, medical records play a much larger role in determining patient similarity. This prioritization results in better model performance, ensuring that our predictions are clinically relevant and useful for healthcare.

\section{Privacy assessment}  \label{sec:aWOE_privacy}
Static data, which consist of patient demographic information and medical history, are sensitive in both legal and ethical contexts. Using these data in machine learning models raises serious privacy concerns, as adversaries often target such information. In the proposed patient similarity model, we performed aWOE technique before using the data in machine learning algorithms. As demonstrated in our previous study \cite{sana2024ppccp}, aWOE serves as a privacy preserving mechanism similar to $k$-anonymity. In particular, it employs a generalization-based $k$-anonymity approach in its transformation process. During the transformation, all values within a given bin are replaced with their corresponding aWOE value, effectively obfuscating the original data. These transformed values are then used to train the machine learning model, making it difficult for adversaries to reconstruct or extract personal information from the model’s outputs. By integrating aWOE, the PSC model inherently enforces privacy guarantees, ensuring data protection throughout its entire pipeline.

\section{Discussion} \label{sec:discussion}
 From the comparison results, it is asserted that the combination of time series data and data transformation based static information has a great impact on improving the patient similarity prediction performance. The previously mentioned work \cite{Mehedy_Masud_2020} studied the effectiveness of the combination of the time series data and static data. However, they did not consider the data transformation methods in the methodology. In this study, we considered two data transformation methods which remarkably improve the prediction performance. The impact of the data transformation methods has been shown for all four different clustering algorithms based PSC techniques. It is well known that DTW based time series similarity which is used in \cite{Vaughan_2016, Franses_2020, brian2020, Donald_1994}, takes a significant amount of computational time. To tackle this issue, we employed a distributed Spark system for the computation of patient similarity. Our experimental results show that the distributed Spark computation reduces computation greatly and meets the demand of real-time patient similarity prediction to help the clinical decision support. Our finding suggests that a healthcare system can use the distributed patient similarity model to find similar patients. The implications of this study offer methodological practices that can significantly enhance both current and future prediction models in the healthcare system. The comparison of our proposed distributed patient similarity models with previous studies and baseline models demonstrates a clear picture of observable performance improvement, as presented in the Section \ref{sec:comparison_previous_study}. Table \ref{table:DPS_F1_score_comparision} reports that the proposed methodology outperforms all the baseline models, demonstrating the supremacy of our proposed approach. The results in Section \ref{sec:effect_of_time_series_data_length} demonstrate that longer time series data significantly improve patient similarity computation, with the 12-hour window offering a practical and effective compromise between performance and feasibility. This insight can guide the development of more accurate and scalable clinical decision support systems tailored to specific diseases and care settings. Additionally, the average execution time taken by the distributed patient similarity technique is 53 seconds (for 100 patients) to determine similarity for a single target patient. Therefore, in a real-world setting, the prediction response time for the proposed DPSC in determining similarity for a single patient would be negligible. Our neighborhood similarity fusion is robust in missing values handling because the corresponding neighborhood will be empty for any missing patient record in a time series variant and will be ignored in the neighborhood fusion during the nearest neighborhood computation. The fusion of time series data and data transformation based static data demonstrates uniform performance improvement as observed from all the experiments. We performed SHAP analysis on our proposed model and discussed the impact of aWOE in the Section \ref{sec:impact_aWOE}. The SHAP analysis indicates that, in our proposed model, patient background or medical records influence similarity prediction more than the patients' demographic information. In conducting patient similarity prediction, privacy concerns are paramount due to the sensitive nature of healthcare. In this study, we also used patients' demographic and medical records, which are highly privacy sensitive. To address the data privacy concerns, we performed a privacy assessment of our proposed models, which demonstrated that the aWOE based model effectively preserves data privacy. We believe the proposed method will be highly beneficial for clinical decision support in patients' medication and treatment recommendations. The contributions of this paper are relevant to both the healthcare industry and academia. Researchers can use the methodology for further research perspectives and the healthcare decision support systems can use it for successful model development.

\section{Conclusions} \label{sec:conclusions}
Patient similarity computation is a critical and potentially life-saving task, especially for emergency patients. In this study, we presented a novel technique that combines time series and static data with distributed computing to find similar patients. Data privacy has always been a serious concern in predicting patient similarity and must be addressed. In this study, we addressed data privacy concerns by performing a privacy assessment of our proposed models, demonstrating the effectiveness of the aWOE-based model in preserving data privacy. Time series patient similarities are computed using the most popular and efficient Dynamic Time Warping (DTW) technique. To reduce the computation time of DTW similarity, a distributed Spark environment has been utilized. The similar patients are selected from their own group of the given patient based on four different clustering algorithms namely, Spectral, K-Means, Agglomerative, and OPTICS. To assess the prediction performance, six widely used measurement metrics have been used such as AUC, accuracy, specificity, precision, recall, and F-measure. This study was conducted on two separate prediction problems: Coronary Artery Disease and Congestive Heart Failure. The proposed DPSC technique has been evaluated on a real dataset (MIMIC-III) and attained commendable performance. The proposed method is empirically compared with a state-of-the-art technique and several baseline models. The experimental outcome indicates that across all measurement metrics, the aWOE based models outperform all other models. To support our findings, the Friedman statistical test and post hoc Holm's test have been performed. All the statistical tests manifest the supremacy of the aWOE based DPSC. The analysis of how varying lengths of time series data affect model performance reveals that a 12-hour observation window is a practical and effective choice for patient similarity computation, particularly in time-critical clinical settings like emergency departments or intensive care units. The comparative analysis showed that our proposed methodology is more effective at identifying similar patients in the context of both the Coronary Artery Disease and Congestive Heart Failure prediction problems and it requires a negligible amount of time to compute, meeting the real-time patient similarity prediction demand. Our next plan is to explore the MIMIC-IV dataset or other real datasets available from hospitals. In addition, we will investigate the effectiveness of our technique on electronic health record (EHR) datasets.

 \section*{Acknowledgments}
 The authors thank Muhammad Abdullah Adnan for his guidance in the Spark computation component of this research.
 %The authors would like to acknowledge that the distributed computation portions of this work were originally presented in the "CSE 6801-Distributed Computing Systems" course at Department of Computer Science and Engineering, Bangladesh University of Engineering and Technology, under the supervision of Muhammad Abdullah Adnan, whose guidance contributed to the development of this research.

%The initial distributed computation portions of this work were originally presented in the "CSE 6801-Distributed Computing Systems" course at Bangladesh University of Engineering and Technology, under the guidance of Muhammad Abdullah Adnan.

%We gratefully acknowledge the foundational work on the Spark computation component of this research, which was initially explored and presented as part of the coursework for CSE 6801: Distributed Computing Systems at Bangladesh University of Engineering and Technology, under the guidance of Muhammad Abdullah Adnan.


%\bibliographystyle{IEEEtran}
%\bibliographystyle{kluwer}
%\bibliographystyle{plainnat}
%\bibliographystyle{ACM-Reference-Format}
%\bibliographystyle{elsarticle-harv}
\bibliographystyle{elsarticle-num}
\bibliography{Bibliography}

 \clearpage

\end{document}